\documentclass[journal,twocolumn]{IEEEtran}
\usepackage{epsfig}
\usepackage{graphicx}
\usepackage{amsmath}
\usepackage{amssymb,subfigure,cases}
\usepackage{setspace}

\usepackage{color,hyperref}
\usepackage{color,soul}
\usepackage{algorithm}
\usepackage{algorithmic}
\usepackage{multirow}
\usepackage{url}

\ifCLASSINFOpdf
\else
\fi

\definecolor{darkblue}{rgb}{0.0,0.0,1.0}
\hypersetup{colorlinks,breaklinks,
	linkcolor=darkblue,urlcolor=darkblue,
	anchorcolor=darkblue,citecolor=darkblue}

\begin{document}
	
	\title{A Joint Convolutional Neural Networks and Context Transfer for Street Scenes Labeling}
	
	\author{Qi~Wang,\IEEEmembership{~Senior Member,~IEEE}, Junyu~Gao, and ~Yuan~Yuan,\IEEEmembership{~Senior Member,~IEEE}
		\thanks{	
			Qi Wang is with the School of Computer Science, with the Unmanned System Research Institute, and with the Center for OPTical IMagery Analysis and Learning, Northwestern Polytechnical University, Xi'an 710072, China (e-mail: crabwq@gmail.com).		
			
			Junyu Gao and Yuan Yuan are with the School of Computer Science and Center for OPTical IMagery Analysis and Learning,
			Northwestern Polytechnical University, Xi'an 710072, Shaanxi, China (e-mail: gjy3035@gmail.com; y.yuan1.ieee@gmail.com).		 		
			
			\copyright 20XX IEEE. Personal use of this material is permitted. Permission from IEEE must be obtained for all other uses, in any current or future media, including reprinting/republishing this material for advertising or promotional	purposes, creating new collective works, for resale or redistribution to servers or lists, or reuse of any copyrighted component of this work in other works.
		}
	}
	\markboth{{IEEE} Transactions on Intelligent Transportation Systems}%
	{Shell \MakeLowercase{\textit{et al.}}: Bare Demo of IEEEtran.cls for Journals}
	\maketitle
	
	\begin{abstract}
		Street scene understanding is an essential task for autonomous driving. One important step towards this direction is scene labeling, which annotates each pixel in the images with a correct class label. Although many approaches have been developed, there are still some weak points. Firstly, many methods are based on the hand-crafted features whose image representation ability is limited. Secondly, they can not label foreground objects accurately due to the dataset bias. Thirdly, in the refinement stage, the traditional Markov Random Filed (MRF) inference is prone to over smoothness. For improving the above problems, this paper proposes a joint method of priori convolutional neural networks at superpixel level (called as ``priori s-CNNs'') and soft restricted context transfer. Our contributions are threefold:
		(1)  A priori s-CNNs model that learns priori location information at superpixel level is proposed to describe various objects discriminatingly;
		(2)  A hierarchical data augmentation method is presented to alleviate dataset bias in the priori s-CNNs training stage, which improves foreground objects labeling significantly;
		(3)  A soft restricted MRF energy function is defined to improve the priori s-CNNs model's labeling performance and reduce the over smoothness at the same time.
		The proposed approach is verified on CamVid dataset (11 classes) and SIFT Flow Street dataset (16 classes) and achieves a competitive performance.
	\end{abstract}


	\IEEEpeerreviewmaketitle

	\section{Introduction}
	\label{Introduction}
	\IEEEPARstart{I}{n} recent years, intelligent driving has been a hot topic for the research communities and industrial companies. It can promote the understanding towards fundamental computer vision and machine learning problems and enhance the actual experience of intelligent transportation.  For this purpose, a critical challenge is how to understand the street scenes  and react to the outside conditions  efficiently. At present, researchers tackle this problem by an integration of several mature technologies, such as pedestrian detection \cite{zhang2015efficient}, anomaly detection \cite{7564410}, vehicle detection \cite{7463502}, road surface detection \cite{wang2015adaptive}, lane detection \cite{7442145} and so on. However, these technologies are on the initial stage of scene understanding and far away from real requirement.

\begin{figure}
		\centering
		\includegraphics[width=.45\textwidth]{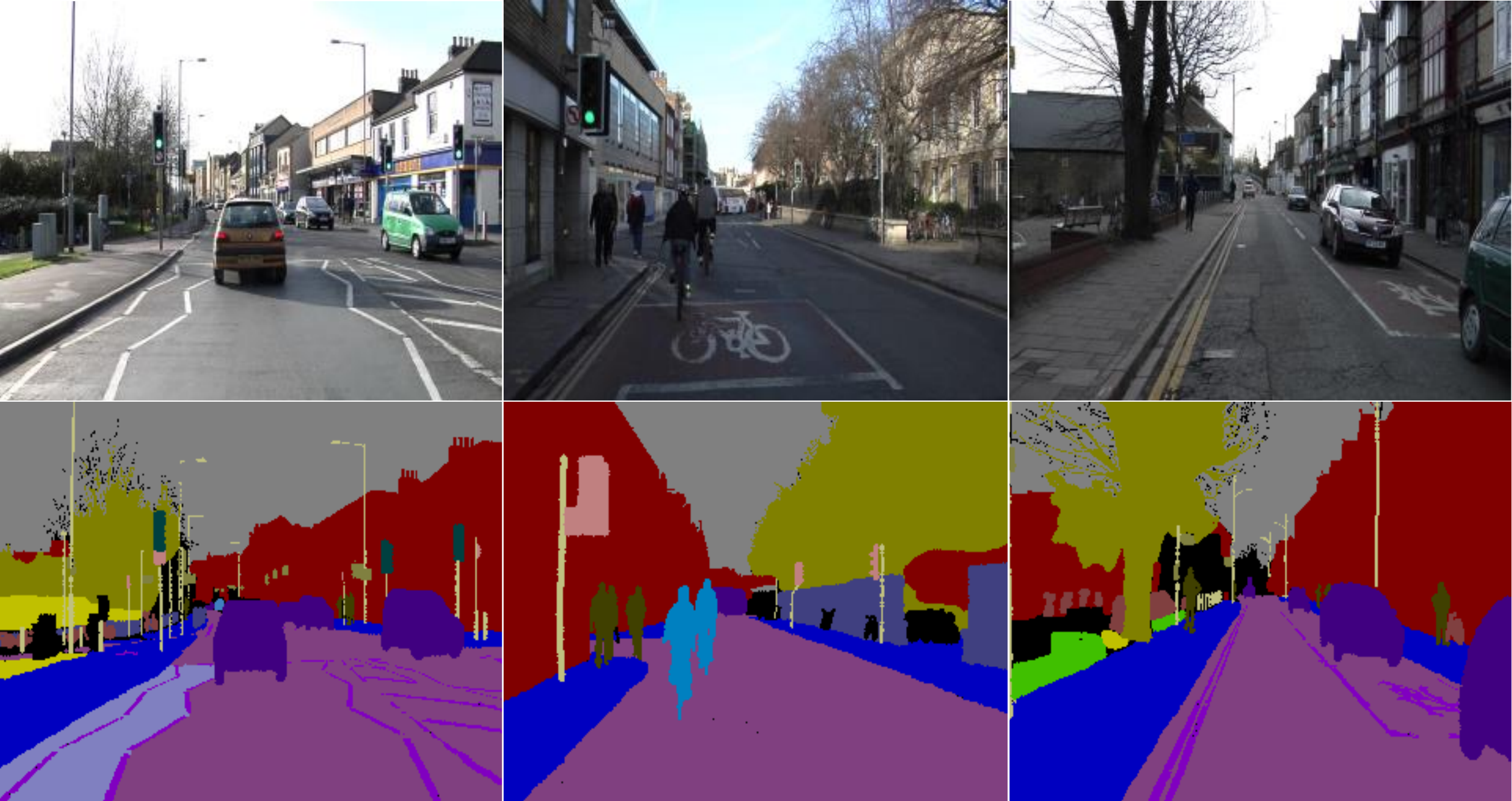}
		\caption{Street scenes labeling examples. The images in the first row are street scenes and the second row illustrates the per-pixel labeling results. }\label{Fig-show}
	\end{figure}

	In order to get a better knowledge of the street scene, a new computer vision task is proposed, semantic scene labeling. It combines segmentation, object detection and multi-object labeling into one single framework and can be regarded as a per-pixel labeling task. This is because for intelligent driving in street scenes, it is necessary to not only recognize the individual participant and event, but also have a thorough perception of the whole view. For instance, if the driver knows where the side buildings are, or what the traffic status is, he will drive more safely. Examples of street scene labeling are presented in Fig. \ref{Fig-show}.

	However, because scene labeling is a unified framework and involves many fundamental computer vision tasks, it is still challenging since Wright \emph{et al.} \cite{DBLP:conf/bmvc/Wright89} firstly put forward this concept in 1989. There are two questions to be solved in this topic: how to get distinctive internal representations of object appearance and how to improve labeling accuracies of foreground objects in the street scenes. Firstly, scene labeling is not like traditional single-object problem that needs to extract features between positive and negative samples. As a multi-object task, how to extract rich and discriminative features to describe different objects is essential to labeling, which is obviously more difficult than single-object task. For this purpose, many approaches (e.g. \cite{nie2016parameter}, \cite{DBLP:conf/eccv/ShottonWRC06}, \cite{DBLP:journals/ijcv/TigheL13},  \cite{7904630}, \cite{brostow2008segmentation}, \cite{sengupta2013urban}) aim to exploit multiple features to characterize objects. The first five exemplar ones compute RGB based features to describe image by combination and fusion of them. The last two exploit 3D features (dense depth maps or 3D point clouds) to reconstruct 3D street scenes. Generally, the more features are extracted, the more information is exploited . However, the feature weight and fusion strategy are manually determined, and it is not easy to obtain some features, especially 3D features. Thus, how to automatically learn rich and discriminative features is an important issue that needs further research.

	At the same time, labeling foreground objects is another intractable issue in the scene labeling. This is because the data distribution in the training set is unbalanced (called as ``long-tail effect''). A few background objects (sky, road and building) account for the majority of the training data, while the foreground objects only take a small part. This phenomenon makes the training process or feature extraction are less adequate and the final result is: background objects labeling accuracy is far higher than foreground objects. However, because foreground objects may be more oriented to intelligent driving than background objects, such as traffic signs, pedestrian, surrounding preceding vehicle and so on, we think the foreground detection is more important to intelligent driving than background detection. Nevertheless, there is no approaches to solve it well because the long-tailed effect is a natural phenomenon and almost exists in every image. If the bias of the dataset can be reduced, the foreground objects detection will be promoted.

	\begin{figure*}
		\centering
		\includegraphics[width=0.95\textwidth]{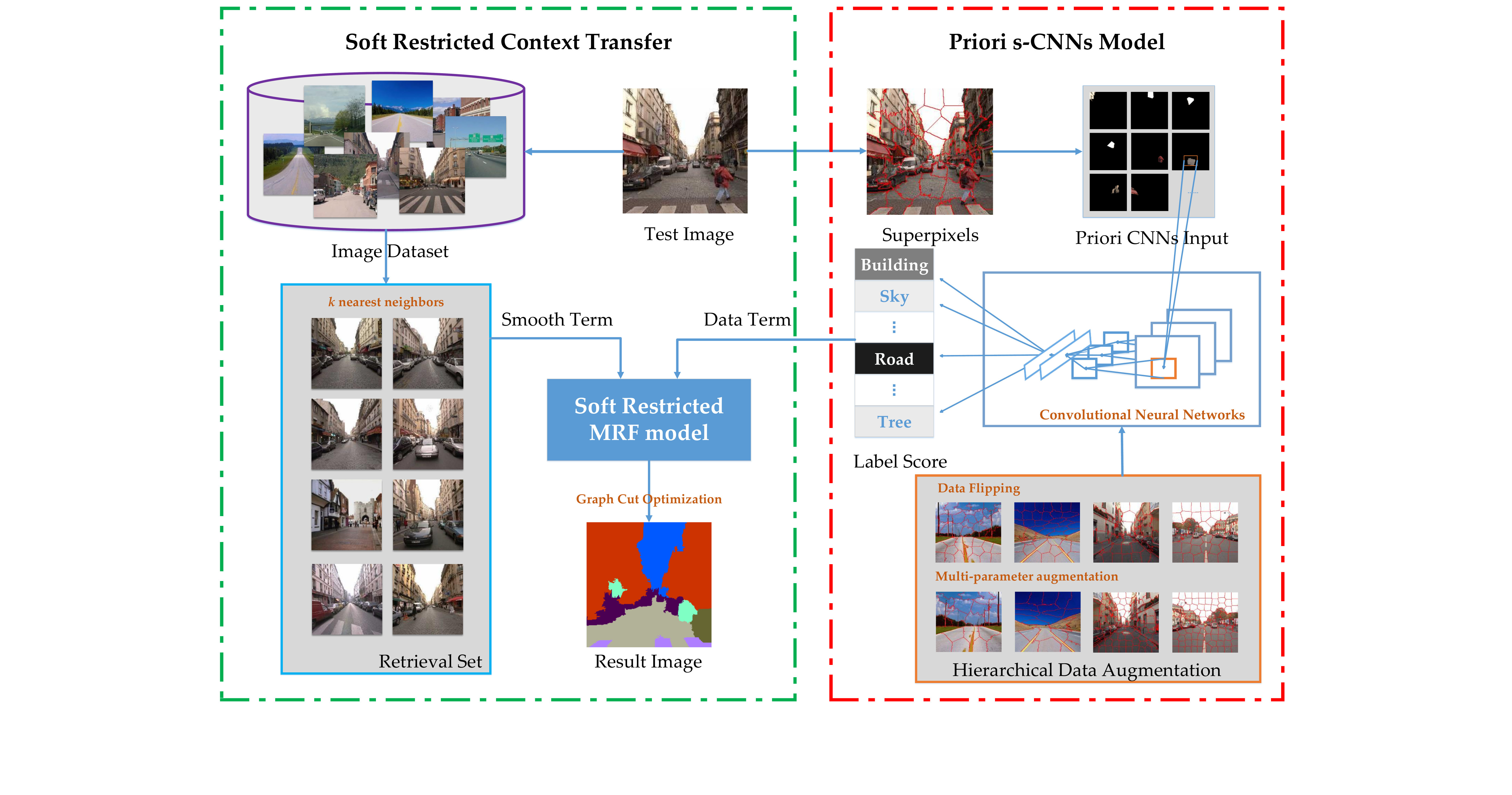}
		\caption{The flowchart of our proposed joint method of priori s-CNNs and soft restricted context transfer. Firstly, given an input image, this paper generates a certain amount of superpixels. For learning priori location information, each superpixel is extracted from original image as a single input of priori s-CNNs. Then, the CNNs model outputs probability vectors (called as ``label score'') corresponding to each label. Secondly, the $k$ nearest neighbors image retrieval searches for similar scenes to test image from the training set by global deep features. After obtaining retrieval set, this work computes conditional probabilities between adjacent superpixels. Finally, a soft restricted MRF model is constructed which integrates label score with priori probability between adjacent superpixels. Through optimizing MRF energy function to refine initial CNNs model's results.}\label{Fig-overview}
	\end{figure*}
	
	Therefore, our model focuses on how to learn more rich features and reduce the bias of dataset for labeling scene more accurately.
	
	\subsection{Overview of Our Approach}
	In this paper, we propose a joint method of superpixel-CNNs model and soft restricted context transfer to tackle the street scene labeling problem. The superpixel-CNNs model focuses on learning rich and discriminative features of image superpixels and exploiting priori location information effectively, which is called as ``priori s-CNNs''. The soft restricted context transfer aims to reduce the noises caused by the priori s-CNNs labeling results. The entire framework is illustrated in Fig. \ref{Fig-overview}.

	\textbf{Training priori s-CNNs with anti-bias data augmentation.} In this stage, a priori s-CNNs is trained to label superpixels. At first, the training images are over-segmented to a certain amount of superpixels. Then all of them are input to the CNNs to train the model parameters by a supervised method. For a more efficient feature learning, the superpixel location prior is particularly considered in this procedure. Thus our CNNs feature does not only contain appearance but also reflect location information. At the same time, in order to reduce the dataset bias, this work uses a hierarchical data augmentation to enlarge the original training set. It takes different numbers of training object classes into account and augments them separately. Thus the CNNs model can learn more rich features to describe superpixels.

	\textbf{Labeling images with context smoothing:} Given a test image, two processes are applied, initial label assignment and context transfer smoothing. The former aims to label each superpixel in the test image according to the previously learned model. But the obtained result is noisy and far from perfect. Therefore, the latter accordingly focuses on reducing the initial labeling noise by transferring contextual clue from the training set to the test image. To this end, we search for the $k$ most similar images in the training set and transfer their structured labels to the examined test image, combined with an MRF post-optimization.

	\subsection{Contributions}
	In this work, we focus on learning more rich features to describe each superpixel and improving the problem of dataset bias. The main contributions of this work are threefold:
	\begin{enumerate}
		\item[1) ]
		Learn rich feature (4096 dimensions) by finetuning the powerful CNNs (AlexNet, an image classifier network) to tackle our task - scene labeling. In order to get a coherent labeling result, we utilize a superpixel with location priors as an input unit instead of a traditional pixel based image. Our treatment keeps the structural relationship between the examined superpixel and the whole image and implicitly embeds the location prior in the CNNs processing. This is critical because the street scene understanding is highly dependent on the class spatial structure. For example, the sky is prone to be in the upper image and the road tends to be in the bottom. Thus, priori s-CNNs can extract rich feature for each superpixel to label scenes.
		\item[2) ]
		Propose a hierarchical data augmentation method to reduce overfitting and dataset bias. Traditional data augmentation expands the training data randomly and equally, which can not balance the number of different training classes. In order to tackle this problem, we propose to enlarge the training set in a more balanced manner. The classes with more training samples will be less augmented, and vice versa. Based on a well adjusted training set, the performance of the foreground objects labeling improve significantly.
		\item[3) ]
		Present a soft restricted MRF model to adapt to priori s-CNNs model's outputs and reduce over smoothness. Traditional approaches treat contributions of adjacent superpixels equally, which causes some foreground objects are smoothed improperly by its majority of adjacent background objects for consistence. In order to weaken this problem, adaptive weights are added to the smoothness term according to the similarity between the pairs. With such a soft restriction, our model can alleviate noises in the initial results and dose not result in serious over smoothness.
	\end{enumerate}

	The rest of the paper is organized as follows. Section \ref{Related Work} reviews the related work briefly. Section \ref{CNNs Model} and Section \ref{context transfer} describe the priori s-CNNs training process and soft restricted context transfer respectively. Section \mbox{\ref{exp}} shows the experimental results on two street scenes datasets and compares its performance with other competitors. In addition, some further discussion and analysis about some import modules in our methods are presented in this section. Finally, we summarize the work in section \ref{conclusion}.
	
	\section{Related Work}
	\label{Related Work}
	In recent years, a large amount of approaches for scene labeling have been proposed.
	According to their pipelines, the algorithms usually consist of two components: extracting image feature and introducing contextual smoothness.

	There are many methods for feature extraction. Liu \emph{et al.} \mbox{\cite{liu2009nonparametric}} use SIFT Flow feature to align the input image. Shotton \emph{et al.} \mbox{\cite{DBLP:conf/eccv/ShottonWRC06}} define a texton and extract its texture, layout, and location information. Tighe and Lazebnik \mbox{\cite{DBLP:journals/ijcv/TigheL13}} compute around 20 features (five types: shape, location, texture, color and appearance) to describe superpixels.  In addition to the above RGB features, some research \mbox{\cite{brostow2008segmentation}}, \mbox{\cite{xiao2009multiple}} and \mbox{\cite{zhang2010semantic}} exploit 3D features, such as 3D point clouds and depth maps. Brostow \emph{et al.} \mbox{\cite{brostow2008segmentation}} propose a method based on 3D point clouds derived from ego-motion. They design five cues (camera height, closest distance to camera, surface orientation, track density and back projection residual) to model patterns of motion and 3D structure. Xiao \emph{et al.} \mbox{\cite{xiao2009multiple}} propose a multi-view parsing method for image sequences. Zhang \emph{et al.} \mbox{\cite{zhang2010semantic}} compute the scene depth information from video sequence through stereo reconstruction of dense depth maps. Peng \emph{et al.} \mbox{\cite{Peng2016:Automatic_full}} propose an unsupervised subspace learning method which can automatically determine the optimal dimension of feature space.
	
	In addition to the above hand-crafted features, deep feature is recently adopted to describe image. Compared to hand-crafted features, it can learn high-level features and fill in representation gap in some way. In an inchoate work, Grangier \mbox{\cite{grangier2009deep}} propose a supervised greedy leaning scheme based on deep convolutional networks. The networks architecture can extract texture, shape, and contextual information. Farabet \emph{et al.} \mbox{\cite{farabet2013learning}} propose a method of learning hierarchical features based on multi-scale convolutional networks. However, because of the lack of training set, this method does not acquire a good results. Until 2014, Girshick \emph{et al.} \mbox{\cite{girshick2014rich}} solve this problem by representation transfer. They propose a region CNN (R-CNN), which use a high-capacity CNNs (AlexNet \mbox{\cite{DBLP:conf/nips/KrizhevskySH12}}) to process region proposal for localizing and segmenting object. Because the AlexNet's parameters are trained on ImageNet dataset, training model based on AlexNet can acquire its robust feature representation. After that, many similar methods exploit this strategy.  Hariharan \emph{et al.} \mbox{\cite{ hariharan2014simultaneous}} aim to detect all instances of a category in an image, and their algorithm is based on region proposals' features by extracting from both the region bounding box  and the region foreground with a jointly trained R-CNN and box CNN. Long \emph{et al.} \mbox{\cite{DBLP:journals/corr/LongSD14}} propose a fully convolutional networks based on AlexNet \mbox{\cite{DBLP:conf/nips/KrizhevskySH12}}, VGG net \mbox{\cite{simonyan2014very}} and GoogLeNet \mbox{\cite{szegedy2014going}}, which only consists of convolutional layers without original fully connected layers.

	As for the contextual information, Markov Random Field (MRF) and Conditional Random Field (CRF) models are very popular solutions. Early methods (e.g. \mbox{\cite{DBLP:conf/eccv/ShottonWRC06}}) exploit local feature information and smoothness prior adjacent pixels by defining second-order potential. Tighe \emph{et al.} \mbox{\cite{DBLP:journals/ijcv/TigheL13}} define a prior conditional probability of adjacent superpixels as contextual information. For exploiting more wide contextual information, Ladicky \emph{et al.} \mbox{\cite{DBLP:conf/iccv/LadickyRKT09}} introduce object detector terms into CRF function. Myeong \emph{et al.} \mbox{\cite{DBLP:conf/cvpr/MyeongCL12, DBLP:conf/cvpr/MyeongL13}} are based on \mbox{\cite{DBLP:journals/ijcv/TigheL13}} and model contextual relationships. Through learning the relationship of superpixels, the scheme transfer the object relationship from retrieved images to test images. Yang \emph{et al.} \mbox{\cite{yang2014context}} incorporate both local and global semantic context information via a feedback based mechanism to refine retrieval set and superpixels matching.

	Besides the above two probabilistic graphical models, context and structure model is also a novel method. Generally speaking, the contextual information propagates in the trees, forests or networks. Sharma \emph{et al.} \cite{sharma2014recursive} propose recursive neural network architecture (contains four networks) for the propagation of contextual information from a superpixel to other one through binary tree. Kontschieder \emph{et al.} \cite{kontschieder2014structured} exploit contextual and structural information in random forests by integrating the structured output predictions into a concise, global, semantic labeling. Long \emph{et al.} \cite{DBLP:journals/corr/LongSD14} integrate appearance representation with semantic information from a shallow and a deep layer respectively. Peng \emph{et al.} \mbox{\cite{Peng2016:Deep_full}} propose a deep subspace clustering methods, which incorporates the structured global prior in representation learning.

	In addition to the above two modules (feature extraction and contextual smoothness), it is important to mention the related works of data augmentation. In the real world, the objects' proportions are imbalanced because of the ``long tail effect''. In the field of knowledge discovery, imbalanced learning is a hot topic, which can affect the performance of learning algorithms in the presence of underrepresented data \mbox{\cite{he2009learning}}. Data augmentation is one of imbalanced learning methods in the deep learning applications. For training AlexNet  \mbox{\cite{DBLP:conf/nips/KrizhevskySH12}}, Krizhevsky \emph{et al.} apply image translation and horizontal reflections. They also alter the intensities of the RGB channels in the training images. Howard  \mbox{\cite{howard2013some}} extends image crops into extra pixels to capture translation and refection invariance, and adds randomly generated lighting which tries to capture invariance to the lighting and minor color variation. Wu \emph{et al.}  \mbox{\cite{DBLP:journals/corr/WuYSDS15}} adopt some color casting, vignetting and lens distortion to augment dataset, which can improve the CNNs' sensitivity to colors that are caused by the illuminants of the scenes.

	\section{Priori s-CNNs based Feature Learning}
	This section mainly explains the training process. Based on a typical CNNs model, we transfer a robust representation to our specific application - scene labeling. For exploiting prior information, we propose a priori based s-CNNs. And for reducing dataset bias and getting a more balanced  model, we propose a hierarchical data augmentation strategy. Thus, our CNNs model can learn rich and discriminative representations to describe images.
	\label{CNNs Model}
	\subsection{Priori s-CNNs}
	\label{psCNNs}
	
	Scene labeling is a task that needs to annotate per pixel, but it is time-consuming to extract features for each pixel and construct a large graph to optimize the MRF energy function. We note that superpixel is a set of pixels that almost belong to the same class and have similar appearance and texture. Once a superpixel is classified as a label, the pixels in the superpixel are assigned as the same label. If we can regard each superpixel as a basic labeling unit, the time cost will decrease significantly. Based on this consideration, we propose a novel superpixel based CNNs, emphasizing the priori effect in the scene labeling application.

	\textbf{Convolutional Neural Networks.} In this paper, in order to label street scenes datasets, we finetune AlexNet  \mbox{\cite{DBLP:conf/nips/KrizhevskySH12}} that is pre-trained on ImageNet Large Scale Visual Recognition Challenge dataset (ILSVRC2012, 1.3 million images, 1000 object categories) by Caffe \footnote{http://caffe.berkeleyvision.org/}. AlexNet consists of 5 convolutional layers and 3 fully connected layers (the last is soft-max layer). For finetuning it, a new soft-max layer replaces the original soft-max layer to predict street scene labeling classes (including the ``void'' class that is representative of the unannotated regions in the datasets).
	
	\textbf{Superpixel Generation.} In this work, superpixel is a basic unit of labeling and each superpixel is produced by a typical and efficient method: simple linear iterative clustering (named as ``SLIC'') \mbox{\cite{achanta2012slic}}. It adopts a k-means clustering algorithm to generate superpixels, which considers the color information in CIE-LAB space and the position of each pixel. SLIC has following advantages: 1) the boundaries of the generated superpixels  are accurately; 2) the generation speed is fast. In a superpixel, nearly all pixels' labels are uniform and belong to the same class. Thus, it is reasonable to regard one superpixel as a processing unit.
	
	\textbf{Priori Superpixel based Processing.}
	\label{superpixels}
	After generating superpixels, we don't resize the them to the same dimension as the input. Alternatively, each superpixel is remained in the original image and the other outside areas are set as black color. Afterwards, these superpixel images enter into the CNNs model and the supervised parameter update is conducted during the training stage. The reason for this operation is explained as follows. In street scenes, we easily find that road region is usually located at the image bottom and the sky on the top. Taking full advantage of this location priori is essential to rule out the false labeling. Therefore, we propose this processing method, which can make CNNs learn location prior of superpixels and more discriminative feature.

	Furthermore, we discuss the effect of location priori in the CNNs. In convolutional layers, because of the parameter sharing and small sizes of convolution kernels ($11 \times 11$, $5 \times 5$ or $3 \times 3$ in AlexNet), the kernel's parameters are not changed by this strategy and they are only sensitive to object class. For example, a feature map $O$ is a 3-D tensor with size of $H \times W \times H$, which is output by a convolutional layer. Here, $H$ is height, $W$ is width and $C$ is the number of channel in the feature map. The $C$-D vector at the $i$-th, $j$-th position in first two dimension represent the appearance information of the corresponding respective field in the input image. In addition, the entire feature map $O$ is viewed as a permutation of the $H \times W$ $C$-D vectors. Such the permutation potentially contains the location information. Then, fully connected layers can integrate the last convolutional layer's feature map into a 4096-dimensional feature vector by inner product operation. Some neurons in these layers are sensitive to the data on the specific channel (the data are output by the specific kernel of the last convolutional layer) of the input. Thus, the fully connected layers model a relationship between appearance features and location priors. In other words, the neurons in fully connected layers can response to specific classes that often appear in specific regions while ignoring other classes. In summary, the fully connected layer can learn location priori for each superpixel.
	
	\begin{figure}
		\centering
		\includegraphics[width=.45\textwidth]{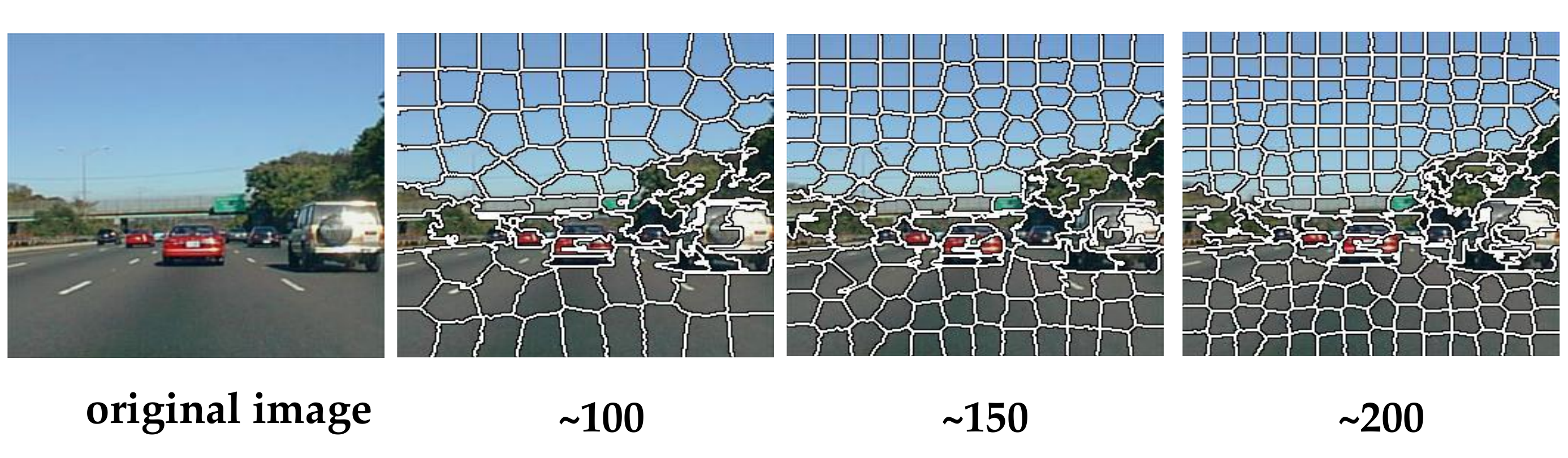}
		\caption{The exemplar display of multi-parameter data augmentation. The number of superpixels is under each image. }\label{multi_scale}
	\end{figure}
	
	\subsection{Hierarchical Data Augmentation}
	In deep learning, overfitting caused by insufficient training data is a common phenomenon. One general method to alleviate overfitting is data augmentation that artificially expands the training set. Traditional strategies include image flipping, rotation, translation, rescaling, shearing, and so on.
	Unfortunately, these strategies share the same characteristic that all training data are expanded randomly and equally. Thus they cannot reduce dataset bias. Besides, some operations, especially rotation and translation, may change the location priori for the vehicle captured video. As a result, designing a self-tailored augmentation method is necessary.
	
	We notice that common street scenes are roughly symmetric in the horizontal direction. Consequently, only horizontal flipping is adopted to avoid location prior changes when enlarging the training data. But this can not solve the dataset bias. In order to get a more balanced training set, different object classes should be augmented distinctively. Based on this consideration, a hierarchical data augmentation mechanism is presented to purposefully enlarge each  class of training set. 
	
	To be specific, the objects in each training image are divided into four categories.
	\begin{enumerate}
		\item[1) ]
		Majority objects: some background objects, such as sky, buildings and roads that count for the most part of an image;
		\item[2) ]
		Common objects: objects with label proportion more than 10\% except ``majority objects'';
		\item[3) ]
		Unusual objects: objects with label proportion more than 3\% and less than 10\%;
		\item[4) ]
		Scarce objects: objects with label proportion less than 3\%.
	\end{enumerate}
	The label proportion is defined as ${{{N_i}} \mathord{\left/
			{\vphantom {{{N_i}} {\sum\limits_j {{N_j}} }}} \right.
			\kern-\nulldelimiterspace} {\sum\limits_j {{N_j}} }}$, where ${N_i}$ is number of pixels labeled as class $i$ in the training image, and $\sum\limits_j {{N_j}}$ is the number of image pixels. All foreground objects labels exist in the last three categories. As for the above four levels, in order to enlarge them to different extents, we present a multi-parameter data augmentation method to generate training data and the concrete implementation is described below.

	In our work, the superpixel is obtained by SLIC \cite{achanta2012slic}. Under various parameters, each training image is over-segmented to different number of superpixels.
	However, not all of them are added to the training set, otherwise the dataset bias cannot be reduced. For complementary augmentation, object with less label proportion will acquire more augmentation with different parameter segmentations. Thus the majority objects get the least segmented superpixels as the training samples, while the scarce objects get the most.
	Eventually, a more balanced training set is achieved.
	
	Actually, this multi-parameter augmentation also has the  multi-scale effect. For example, in Fig. \mbox{\ref{multi_scale}}  the same image is segmented by different parameters of SLIC and some foreground objects with distinctive appearances and sizes are all involved in the training set. This makes the model learn more diverse and rich features.
	
	\begin{figure}
		\centering
		\includegraphics[width=.45\textwidth]{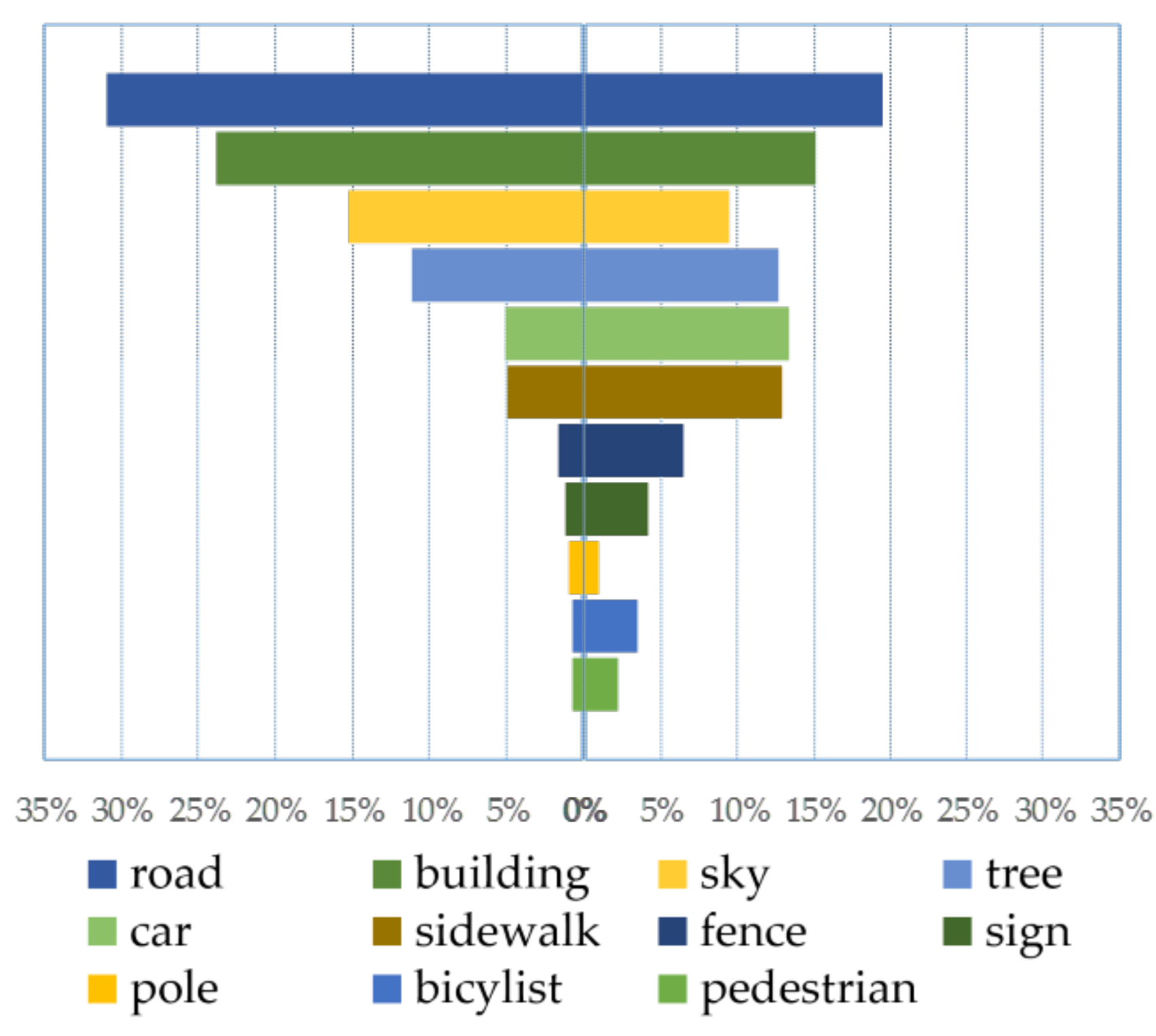}
		
		\caption{The label distribution histogram of CamVid dataset \cite{brostow2008segmentation}. The left is the label distribution of the original dataset before data augmentation, which shows a ``Long Tail Effect''. The right demonstrates a distribution after hierarchical data augmentation, which is more balanced among the different classes.}\label{augmentation}
	\end{figure}

	In order to illustrate the effect of this self-tailored augmentation, we  take CamVid \cite{brostow2008segmentation} dataset as an example. The label  distribution histograms  before and after data augmentation are shown respectively in Fig. \ref{augmentation}. The left part demonstrates clearly the ``long tail effect" before augmentation. As can be seen from the bars, some background classes (sky, building and road) account for more than 70\% of all training data, but the important foreground objects (person, pole, fence, sign, etc.) only cover a few proportion. According to the proposed augmentation strategy, we divide these objects into four categories and expand their numbers through horizontal flipping and multi-parameter segmentation. The resulting proportion histograms alleviate the ``long tail effect" greatly, which can be seen in the right part of Fig. \ref{augmentation}. According to statistics, the number of superpixels in the training set increases from around 60,000 to more than 130,000 and the proportions of common, unusual and scarce objects raise greatly compared to the original training set.  Therefore, we can say a more balanced training set is obtained after data augmentation.\label{pole}
	
	\subsection{Local Superpixel Labeling}
	\label{initial_result}
	With the above treatment, we can train the priori s-CNNs effectively. Then given a test image, the soft-max layer outputs a label score vector $s$ for each superpixel, which represents the probability of being labeled as each class. Selecting the label with largest score as the superpixel's label and combining all the labels of superpixels in one test image forms the initial labeling result. However, only exploiting local feature is not enough, because the results may be noisy. Thus the initial results should be refined further.
	
	\section{Soft Restricted Context Transfer}
	\label{context transfer}
	In Section \mbox{\ref{initial_result}}, the initial labeling results are obtained. Nonetheless, since the superpixels are individually examined, the spatial coherence needs to be improved further. SuperParsing  \mbox{\cite{DBLP:journals/ijcv/TigheL13}} propose an effective method to utilize contextual information. It comprises two steps, nearest image retrieval and MRF optimization. However, the nearest image retrieval adopts hand-crafted features, which can not represent high-level image information. What's more, the traditional MRF model can result in over-smoothness. For solving the questions, we exploit deep features to search neighbors and propose a soft restricted MRF model, which can utilize internal difference in adjacent superpixels to alleviate the over-smoothness.

	\subsection{The $k$ Nearest Image Retrieval}
	In order to transfer more accurate contextual information to the test image, the scenes that contain similar content structure should be considered. Thus we retrieve the $k$ nearest images in the training set for the examined test image and exploit their contextual influence.
	
	This system computes a deep global image features to search neighbors, which is 4096-D vector from fc7 layer of AlexNet. Here, the AlexNet is trained on Places Database (scene-centric databases, more than 7 million images, including 205 scenes categories), named as Places-CNN \mbox{\cite{zhou2014learning}}. The model can effectively extract global feature of scenes. For each image in the training set, it will be ranked according to the increasing order of Euclidean distance to the test image on the computed 4096-D deep feature. Then the nearest $k$ neighbors of the test image are chosen to transfer the contextual information in the next step (soft restricted MRF model inference). Compared to some traditional methods such as SuperParsing \mbox{\cite{DBLP:journals/ijcv/TigheL13}}, deep features describe appearance and high-level semantic information better than hand-crafted features (More discussions about the advantages of deep features are presented in Section \mbox{\ref{compare_features}}). Thus, the more accurate contextual information is computed and transferred by Soft Restricted MRF Model in the next section.

	\subsection{Soft Restricted MRF Model Inference}
	\label{MRF}
	For transferring contextual information from the retrieval set, the MRF model is a popular method. Given a superpixel, traditional methods employ its adjacent superpixels to smooth  it equally. However, this is not a reasonable strategy. We think a pair of  similar superpixels should smooth each other more than dissimilar pairs. Thus, we propose a soft restricted MRF model, which weights each adjacent superpixel to measure its contribution of spatial coherence.
	
	We formulate an MRF energy function over the field of superpixel labels ${\bf{l}} = \left\{ {{l_i}} \right\}$ as:
	\begin{equation}
	\begin{array}{l}
	E({\bf{l}}) = \sum\limits_{{s_i} \in SP} {{E_d}\left( {{s_i},{l_i}} \right)}  + \lambda \sum\limits_{\left( {{s_i},{s_j}} \right) \in {\varepsilon _w}} {{E_s}\left( {{l_i},{l_j}} \right)} ,
	\end{array}
	\end{equation}
	where $SP$ is the set of superpixels in the test image, superpixel ${s_j}$ is adjacent to superpixel $ {s_i}$, ${{\varepsilon _w}}$ is the set of edges of adjacent superpixels, and $\lambda $ is a smoothing constant. The data term ${E_d}\left( {{s_i},{l_i}} \right)$ denotes the cost of assigning superpixel $s_i$ with label ${{l_i}}$ and the  definition is:
	\begin{equation}
	\begin{array}{l}
	{E_d}\left( {{s_i},{l_i}} \right) = {\left( {A_{{s_i}}^s - A_{{s_i}}^r\left( {{l_i}} \right)} \right)^2},
	\end{array}
	\end{equation}
	where ${A_{{s_i}}^s}$ is the output label score vector for superpixel $s_i$ from the priori s-CNNs, ${A_{{s_i}}^r\left( {{l_i}} \right)}$ is the observation value, an indicator vector whose ${l_i}$-th item is set as $1$ and others $0$. Suppose that the $p$-th item of label score vector have largest probability, if ${l_i} = p$, the ${E_d}\left( {{s_i},{l_i}} \right)$ will be smallest and vice versa. The smoothness term ${E_s}(l_i,l_j)$ stands for the cost of a suerpixel smoothed by adjacent superpixels. If a pair of adjacent superpixels appear in the retrieval set frequently, the smoothness term should be small. This term is defined based on probabilities of label co-occurrence statistics:
	\begin{equation}
	\begin{array}{l}
	{E_s}\left( {{l_i},{l_j}} \right) =  - {w_{ij}} \times \log \left[ {\dfrac{{P\left( {{l_i}|{l_l}} \right) + P\left( {{l_j}|{l_i}} \right)}}{2}} \right] \times \delta \left[ {{l_i} \ne {l_j}} \right],
	\end{array}
	\end{equation}
	where ${P({l_i}|{l_j})}$ is the conditional probability of assigning label ${l_i}$ to the superpixel given its neighbor has label ${l_j}$, which is estimated from its corresponding retrieval set. ${w_{i,j}}$ is a soft restriction on the smoothness term, which represents the contribution of each pair of adjacent superpixels. It is defined as the squared Euclidean distance between label scores of two adjacent superpixels:
	\begin{equation}
	\begin{array}{l}
	{w_{ij}} = {\left( {A_{{s_i}}^s - A_{{s_j}}^s} \right)^2},
	\end{array}
	\end{equation}
	where ${A_{{s_i}}^s}$ and ${A_{{s_j}}^s}$ denote the label score of superpixel $s_i$ and $s_j$. The more alike the adjacent superpixels are, the smaller ${w_{ij}}$ is, and vice versa. Thus, ${w_{ij}}$ enhances the smoothness between similar superpixels and reduces the smoothness between distinguished superpixels. The last factor $\delta \left[ {{l_i} \ne {l_j}} \right]$ can be considered as a Potts penalty, which is necessary to ensure that this term is semi-metric  \cite{DBLP:journals/pami/BoykovVZ01}. It is defined as below:
	\begin{equation}
	\begin{array}{l}
	\delta \left[ {{l_i} \ne {l_j}} \right] = \left\{ \begin{array}{l}
	0\quad \quad if\;{l_i} = {l_j}\\
	1\quad \quad if\;{l_i} \ne {l_j}
	\end{array} \right.,
	\end{array}
	\end{equation}
	If the assigned labels for the two adjacent superpixels are the same, the smoothness term is not necessary and should be set as $0$.
	
	Exploiting the prior conditional probability aims to reduce labeling errors. For example, if a superpixel is a part of a person,  it may be assigned with a label ``pedestrian'' or ``bicyclist''. But if its   adjacent superpixels are likely to be ``sidewalk'', it is  more probable to label it with ``pedestrian'' according to the learned prior conditional probability from the retrieval set.
	We perform MRF inference using an efficient graph cut optimization\footnote{The C++ code and MATLAB wrapper are developed by O. Veksler and A. Delong and available at http://vision.csd.uwo.ca/code/gco-v3.0.zip}\cite{DBLP:journals/pami/BoykovVZ01, kolmogorov2004energy, boykov2004experimental}.

	\section{Experiment}
	In this section, we report experimental details and results on the two challenging datasets: CamVid \mbox{\cite{brostow2008segmentation}} and SIFT Flow Street dataset. Section \mbox{\ref{metric}}  shows the two evaluation criteria in scene labeling. Section \mbox{\ref{dataset}} presents some details and characteristic of the two datasets. Section \mbox{\ref{setup}} gives parameter setup and implementation details in the experiments. Then, the results and discussions are presented in Section \mbox{\ref{cam_exp}} and \mbox{\ref{sift_exp}}. Finally, we discuss the effects of the proposed priori location, the advantages of the soft restricted MRF model, the comparison between CNN and hand-crafted features in image retrieval and convergence issues of the stepwise models in last four sections (\mbox{\ref{Prior_s-CNNs}}, \mbox{\ref{soft_MRF}}, \mbox{\ref{compare_features}} and \mbox{\ref{convergence}}).
	\label{exp}

	\subsection{Evaluation Criteria}
	\label{metric}
	In the scene labeling field, there are two metrics to evaluate each algorithm's performance: per-pixel accuracy and mean-class accuracy. The former is defined as:
	\begin{equation}
	\begin{array}{l}
	{r_p} = \dfrac{{\sum\nolimits_i {{n_{ii}}} }}{{\sum\nolimits_i {\sum\nolimits_j {{n_{ij}}} } }},
	\end{array}
	\end{equation}
	where ${{n_{ij}}}$ is the number of pixels  assigning label $i$ as label $j$, ${\sum\nolimits_i {\sum\nolimits_j {{n_{ij}}} } }$ and ${\sum\nolimits_i {{n_{ii}}} }$ stand for the total number of pixels  and total number of pixels that are assigned correct label, respectively. However, because the label distribution suffers from unbalanced problem in practice, only adopting the per-pixel accuracy is not precise. Moreover, in the street scenes, the foreground objects are essential to safe driving, but their contribution to per-pixel accuracy is limited. Therefore, a more reasonable criterion should be introduced. Specifically, the mean-class accuracy is defined as below:
	\begin{equation}
	\begin{array}{l}
	{r_c} = \dfrac{1}{N}\sum\nolimits_i {\dfrac{{{n_{ii}}}}{{\sum\nolimits_j {{n_{ij}}} }}} ,
	\end{array}
	\end{equation}
	where $N$ denotes the number of the label classes. It is an average of per-pixel accuracy of each class, so it can evaluate the overall performance at the class level.
	\subsection{Dataset}
	\label{dataset}
	\subsubsection{CamVid Dataset}
	The Cambridge-driving Labeled Video Database (CamVid)\footnote{http://mi.eng.cam.ac.uk/research/projects/VideoRec/CamVid/} is a challenging road driving scenes dataset, which includes 4 video sequences (one video is divided into 2 parts) with image size of $960 \times 720$ pixels. Similar to \cite{DBLP:journals/ijcv/TigheL13}, \cite{yang2012local} and \cite{kontschieder2014structured},
	we merge the 32 object classes of the original dataset into 11 classes. They are road, building, sky, car, sign-symbol, tree, pedestrian, fence, column-pole, sidewalk and bicyclist. Table \ref{camvid_table} shows the detailed information of CamVid dataset.
	\begin{table}[htbp]
		
		\centering
		\caption{The detail information of CamVid dataset is shown as below, including sequence name, the number of frames, data type and scene category. }
		\label{camvid_table}
		\begin{tabular}{ccccccc}
			\hline
			Video sequence &Frame no. &Type &Scene \\
			\hline
			0001TP-1   &62  &train &dusk \\
			0016E5    &305  &train &daytime \\
			0006R0   &101  &train &daytime  \\
			\hline
			0001TP-2  &62 &test & dusk \\
			Seq05VD   &171 &test &daytime  \\
			\hline
		\end{tabular}\label{tableshow}
	\end{table}
	
	\subsubsection{SIFT Flow Street Dataset}
	
	The original SIFT Flow dataset\footnote{1http://people.csail.mit.edu/celiu/LabelTransfer/LabelTransfer.rar} consists of 2,688 images of 33-class outdoor scenes, which is selected from LabeleME \mbox{\cite{russell2008labelme}} and annotated by LabelME's users. These outdoor scenes include coast, forest, highway, inside city, street scenes and so on, with a resolution of $256 \times 256$. For doing the experiments in the specific street context, we only choose a part of them as our dataset which is called ``SIFT Flow Street Dataset''.

	To be specific, we select the highway and street scenes from SIFT Flow dataset and remove those images that are not from the perspective of vehicles. The new dataset consists of 529 images (491 training images and 38 testing images are selected from original training and testing sets respectively). At the same time, the original label classes are updated by removing the unrelated labels.  Eventually, there are 16 classes (road, sky, sidewalk, building, tree, car, field, fence, person, crosswalk, sign, streetlight, bus, bridge, window, and mountain) in the SIFT Flow Street dataset.
	
	\subsection{Implementation Details \& Settings}
	\label{setup}
	
	\textbf{Settings of the priors s-CNNs.} As for each image (training or testing), it is resized to $256 \times 256$px to adapt to the CNNs model and is oversegmented to $\sim 150$ superpixels (we treat ``$\sim 150$'' as ``the main parameter''). In the training priori s-CNNs stage, the learning rate is initialized at $10^{-4}$ and reduced ten times every ten thousand iterations. Our models are only sensitive to the learning rate: the smaller value selection results in the slower convergence speed and the higher loss. On the contrary, setting the more larger learning rate does not makes the model converge. 
	
	\begin{figure}
		\centering  
		\subfigure[CamVid dataset.] { \label{fig_para_cam} 
			\includegraphics[width=0.95 \columnwidth]{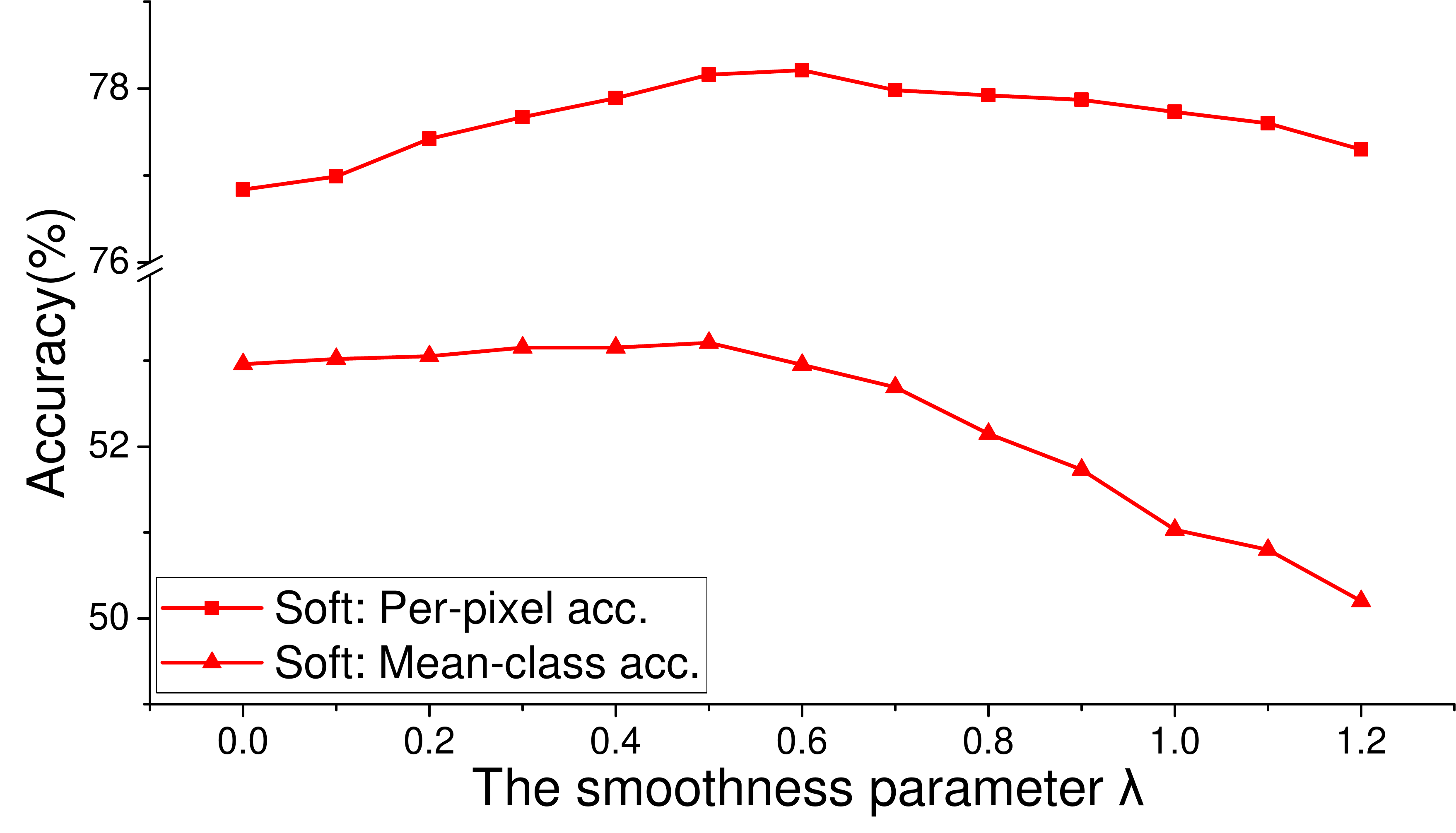} }    
		\subfigure[SIFT Flow Street dataset.] { \label{fig_para_sift}    
			\includegraphics[width=0.95 \columnwidth]{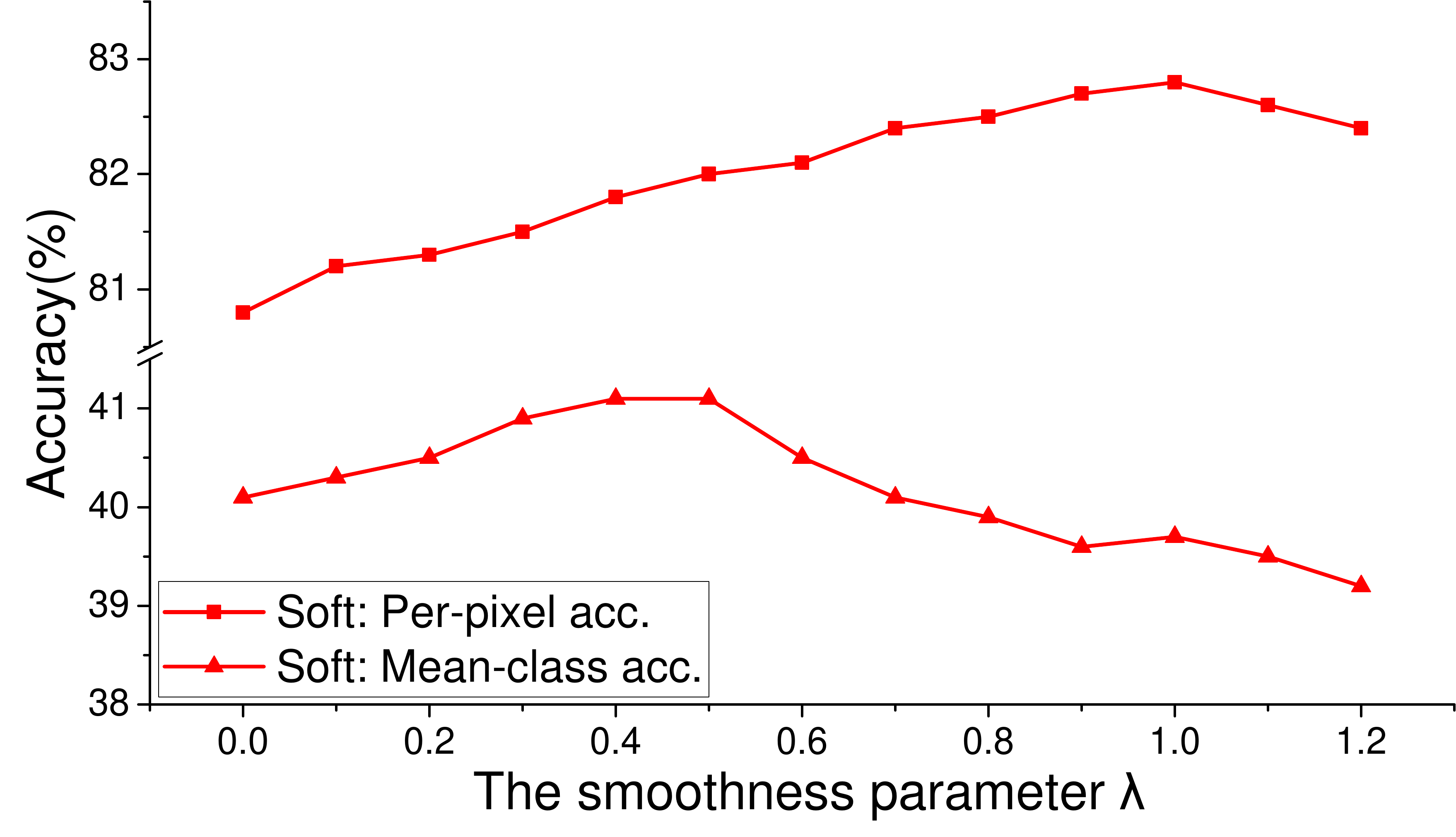} }    
		\caption{The red solid lines demonstrate the effects of our proposed soft restricted MRF model under the different smoothness parameter $\lambda$. The (a) is on CamVid dataset, and the (b) is on SIFT Flow Street dataset.}    \label{para_all}  
	\end{figure}

	\textbf{Settings of the hierarchical data augmentation.} As for the data augmentation, the majority objects are not enlarged; the other training samples are horizontally flipped but they are segmented by different parameters. Specifically, the common objects are expanded  under the main parameter; the unusual objects are augmented under the parameters of $100$, $125$ and $200$. In addition to the above parameters, the scarce objects are expanded under more parameters, including $175$, $130$ and $170$. With this strategy, a more balanced training set is obtained.
	
	\textbf{Settings of the context transfer.} In the $k$ nearest image retrieval,  $k$ is set as a default value $50$ \cite{DBLP:journals/ijcv/TigheL13}, which can achieve  the best mean-class accuracy. Another important parameter is $\lambda$ in the soft restricted MRF model. Fig. \ref{para_all} demonstrates the performance under different $\lambda$ choices on the two datasets.   As can be seen from the red lines, the mean-class accuracy is almost stable at the beginning and decreases with the increase of $\lambda$. The per-pixel accuracy   increases firstly and then decreases. This is because with the increase of $\lambda$, the labels with small area (e.g., foreground objects) are over-smoothed. Therefore, a moderate $\lambda$ might be appropriate. Since it is more important to obtain a high mean-class accuracy than per-pixel accuracy for preserving the foreground objects, $\lambda$ is set to $0.5$ in this work.

	After setting the above parameters, the entire model will perform automatically and without manual operation.

	\textbf{Settings of the compared algorithms.} For showing the superiority of our method, the five mainstream algorithms are added to the comparison. They are SuperParing \mbox{\cite{DBLP:journals/ijcv/TigheL13}}, LLD \mbox{\cite{yang2012local}}, LOR \mbox{\cite{DBLP:conf/cvpr/MyeongCL12}}, SLiRF \mbox{\cite{kontschieder2014structured}}, THSRT \mbox{\cite{DBLP:conf/cvpr/MyeongL13}} and FCN \mbox{\cite{DBLP:journals/corr/LongSD14}}. The first five traditional approaches all exploit hand-crafted features: SuperParsing, LOR and THSRT use 20 features to represent superpixels; LLD designs a local label descriptor by concatenating label histogram; and SLiRF exploits low-level image features. The last two, FCN-32s and FCN-8s \mbox{\cite{DBLP:journals/corr/LongSD14}} that are finetuned on AlexNet, exploit the fully convolutional network to labeling scenes end-to-end. Because of no source code, we do not test LLD \mbox{\cite{yang2012local}} and SLiRF \mbox{\cite{kontschieder2014structured}} on SIFT Flow Street dataset.

	\subsection{Performance on CamVid Dataset}
	\label{cam_exp}
	
	\begin{figure*}
		\centering
		\includegraphics[width=.95\textwidth]{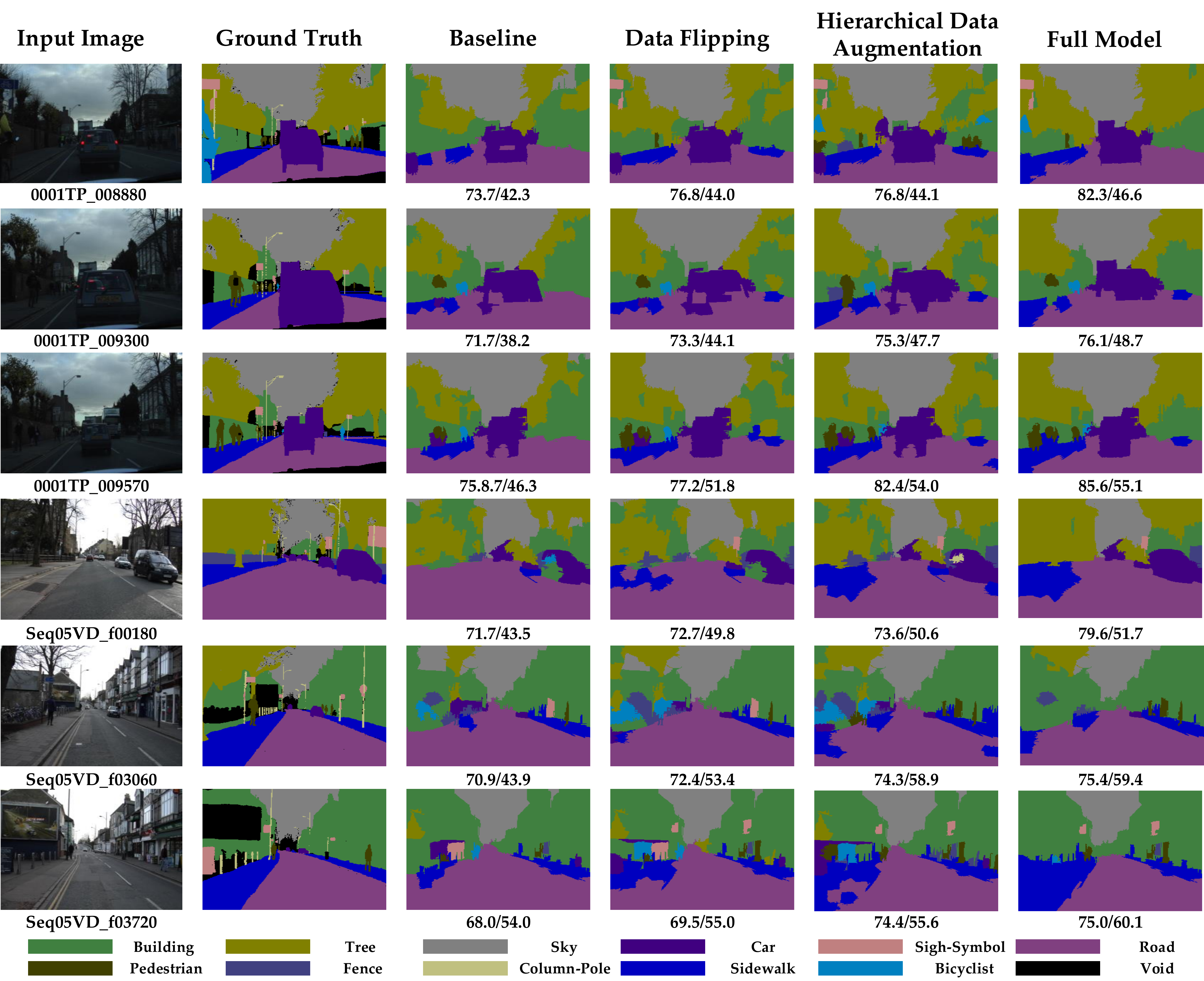}
		\caption{Exemplar results on CamVid dataset. We report the four comparative results, namely the baseline, baseline$+$data flipping, baseline$+$hierarchical data augmentation and full model (baseline$+$hierarchical data augmentation$+$soft restricted MRF inference), respectively. The values  under each image are the per-pixel/mean-class accuracies. The test images of the first three rows are from dusk sequence ``0001TP'' and the others are selected from the daytime sequence ``Seq05VD''.}\label{cambid_show}
	\end{figure*}
	
	Table \ref{cam_table_overall} shows the two metrics of different comparative methods. At first, the baseline only uses the original data to train CNNs model. Then the traditional data flipping and our hierarchical data augmentation are added to the training process respectively. The last one is the soft restricted MRF inference based on the CNNs model with the hierarchical data augmentation (called as ``full model'').

	\begin{table}[htbp]
		\centering
		\caption{Comparison of different approaches on CamVid dataset}
		
		\begin{tabular}{lcccccc}
			\hline
			Methods &Per-pixel &Mean-class \\
			\hline
			SuperPasing \cite{DBLP:journals/ijcv/TigheL13}(Still Image) &78.6\% &43.8\% \\
			LLD \cite{yang2012local} &73.7\% &36.6\% \\
			LOR \cite{DBLP:conf/cvpr/MyeongCL12}  &72.5\%  &35.7\%  \\
			THSRT \cite{DBLP:conf/cvpr/MyeongL13}  &73.1\%  &35.7\%  \\
			SLiRF \cite{kontschieder2014structured}  &72.5\%  &51.4\%  \\
			FCN-32s (AlexNet) \mbox{\cite{DBLP:journals/corr/LongSD14}}  &80.1\% &44.7\%\\
			FCN-8s (AlexNet) \mbox{\cite{DBLP:journals/corr/LongSD14}} &\textbf{80.8\%} &47.4\%\\
			\hline
			\textbf{Ours methods:}\\
			Baseline   &77.1\%   &45.6\%  \\
			Data Flipping &77.2\%  &47.9\%  \\
			Hierarchical Augmentation  &76.8\%  &53.0\%  \\
			Full Model  &78.1\%  &\textbf{53.2\%}  \\
			\hline
			
		\end{tabular}\label{cam_table_overall}
	\end{table}

	It can be seen that our per-pixel accuracy of 78.1\% is not the best, but for mean-class accuracy, our full model achieves the best performance. Considering the above criteria collectively, our results is the best in all of the methods. On the one hand, compared with the traditional strategies, our priori s-CNNs can learn more discriminative features to describe various objects. Thus, more foreground objects are labeled accurately. On the other hand, compared with the FCN-$x$s \mbox{\cite{DBLP:journals/corr/LongSD14}}, our model labels the foreground objects more accurately.

	Next, we discuss the effects of hierarchical data augmentation. Our proposed augmentation method can improve the mean-class accuracy (from 45.6\% to 53.2\%, increasing by 16.7\%) more significantly than traditional data flipping (from 45.6\% to 47.9\%, increasing by 5.0\%). But for the per-pixel accuracy, the improvement is not obvious. The reason is that the labeling performance of foreground objects increases but the background objects' drops simultaneously. More discussions will be presented in the next paragraph.

	In order to analyze the labeling performance further, the results of each class are shown in Table \ref{camvid_table_show}. According to the distribution of bold statistics, we find that the best performance  of almost all   foreground classes are in the bottom two rows which utilize  the hierarchical data augmentation. However, there is an exception, namely, ``column-pole'', whose performance is not good (2.2\% versus the best 4.1\% \cite{yang2012local}). The reason is that the ``column-pole'' training data can not be segmented perfectly by SLIC and is not enlarged effectively (as shown in Fig. \ref{augmentation}).
	We also notice that  the accuracy of the majority objects (sky, building and road) decreases slightly after data augmentation. This is because with the increase of foreground labeling,  the boundary pixels that previously belong to the background change their labels due to the unprecise superpixel segmentation.

	\begin{table*}[htbp]
		\newcommand{\tabincell}[2]{\begin{tabular}{@{}#1@{}}#2\end{tabular}}
		\centering
		\caption{Comparison of per-class accuracy with SuperPasing \cite{DBLP:journals/ijcv/TigheL13}, LLD \cite{yang2012local}, LOR \cite{DBLP:conf/cvpr/MyeongCL12} and FCN \mbox{\cite{DBLP:journals/corr/LongSD14}} on CamVid dataset. }\label{camvid_result_table}
		\begin{tabular*}{1\textwidth}{@{\extracolsep{\fill}}  |l|c c c c c c c c c c c|}
			\hline
			&\rotatebox{90}{Building} & \rotatebox{90}{Tree} & \rotatebox{90}{Sky} & \rotatebox{90}{Car} & \rotatebox{90}{Sign-Symbol} & \rotatebox{90}{Road} &\rotatebox{90}{Pedestrian} & \rotatebox{90}{Fence} & \rotatebox{90}{Column-Pole} & \rotatebox{90}{Sidewalk} &\rotatebox{90}{Bicyclist}\\     
			\hline
			SuperPasing \cite{DBLP:journals/ijcv/TigheL13}(Still Image) &84.8&65.1&94.7&47.5&24.6&96.2&8.3&9.1&3.4&43.7&3.9  \\   
			LLD \cite{yang2012local}&80.7&61.5&88.9&16.4&-&\textbf{98.0}&1.1&0.01&\textbf{4.1}&12.5&0.01 \\
			LOR \cite{DBLP:conf/cvpr/MyeongCL12}&84.3&29.4&93.1&45.6&1.0&94.0&1.3&0.5&1.3&39.5&2.6 \\
			THSRT \cite{DBLP:conf/cvpr/MyeongL13}&\textbf{87.2} &27.7&91.9&43.2&0.4&93.9&1.4&0.03&0.4&43.4&3.1 \\
			FCN-32s (AlexNet) \mbox{\cite{DBLP:journals/corr/LongSD14}} &85.5 &63.6&90.3&63.4&10.4&94.1&5.0&10.7&0.0&69.0&0.3\\
			FCN-8s (AlexNet) \mbox{\cite{DBLP:journals/corr/LongSD14}} &82.3&67.8&92.2&66.0&15.3&94.2&7.1&22.0&0.1&\textbf{71.8}&2.6\\
			\hline
			\textbf{Our methods:} & & & & & & & & & & &\\
			Baseline &84.9 &60.8 &\textbf{95.3} &63.7 &22.0 &96.2 &24.0 &15.0 &1.0 &23.3 &15.0  \\
			Data Flipping &79.1 &70.0 &94.2 &67.4 &26.6 &95.2 &28.3 &16.9 &2.1 &30.5 &17.0 \\
			Hierarchical Augmentation &70.4 &73.9 &93.6 &68.8 &\textbf{31.0} &92.4 &\textbf{38.9} &\textbf{32.7} &2.3 &50.3 &28.4  \\
			Full Model &74.4 &\textbf{74.9} &93.8 &\textbf{69.8} &29.6 &92.6 &38.5 &29.8 &2.2 &52.0 &\textbf{29.0} \\
			\hline
		\end{tabular*}\label{camvid_table_show}
	\end{table*}

	For reporting the advantages of the our algorithm, Fig. \ref{cambid_show} shows six typical exemplar labeling results. At first, we show the impacts of the hierarchical data augmentation on the labeling results. Without the data augmentation, the ``tree'' and ``building'' are prone to be mixed. After the hierarchical data augmentation, they are distinguished more clearly. In addition, more other foreground objects (sidewalk, pedestrian, sign and so on) are also labeled. For example, in the first input image, by our data augmentation the two signs are labeled correctly; in the second, third and last input images, several persons are labeled as ``pedestrian'' after data augmentation. Secondly, we present the effects of soft restricted context transfer. In the forth input image, the parts of the car are mislabeled as building, bicyclist and column-pole without the soft restricted MRF model inference. After considering contextual information by the MRF model, the car can be labeled entirely and accurately. In the fifth image, the left building is recognized as fence, bicyclist, car, pedestrian and so on. After smoothing this result, the error is mitigated considerably.

	\subsection{Performance on SIFT Flow Street Dataset}
	\label{sift_exp}
	The results of SuperParsing \mbox{\cite{DBLP:journals/ijcv/TigheL13}}, \mbox{LOR \cite{DBLP:conf/cvpr/MyeongCL12}}, THSRT \mbox{\cite{DBLP:conf/cvpr/MyeongL13}}, FCN \mbox{\cite{DBLP:journals/corr/LongSD14}} and our models are listed in Table \mbox{\ref{sift_table}}. From the table, we can see our full model achieves an excellent result (82.0\% per-pixel accuracy and 41.1\% mean-class accuracy). Obviously, our mean-class accuracy of 41.2\% outperforms the SuperParsing (32.8\%)\cite{DBLP:journals/ijcv/TigheL13}, LOR (34.2\%)\cite{DBLP:conf/cvpr/MyeongCL12}, THSRT (33.2\%) \cite{DBLP:conf/cvpr/MyeongL13} and FCN (36.8\% and 37.2\%) \mbox{\cite{DBLP:journals/corr/LongSD14}}. Compared with these mainstream methods, our model is trained on a more balanced dataset, which can learn rich and discriminative features to describe various objects. Thus, our model achieves the best mean-class accuracy. However, our per-pixel accuracy is not the best but it is close to the best (best 84.7\% \mbox{\cite{DBLP:journals/corr/LongSD14}}). As a whole, our performance is competitive on the two criteria compared to other popular methods.

	\begin{table}[htbp]
		
		\centering
		\caption{Comparison of different approaches on SIFT Flow Street dataset.}
		
		\begin{tabular}{lcccccc}
			\hline
			Methods &Per-pixel &Mean-class \\
			\hline
			SuperParsing \cite{DBLP:journals/ijcv/TigheL13} &79.9\% &32.8\% \\
			LOR \cite{DBLP:conf/cvpr/MyeongCL12} &84.3\% &34.2\% \\
			THSRT \cite{DBLP:conf/cvpr/MyeongL13}  &83.7\%  &33.2\%  \\
			FCN-32s (AlexNet) \mbox{\cite{DBLP:journals/corr/LongSD14}} &84.0\%  &36.8\%\\
			FCN-8s (AlexNet) \mbox{\cite{DBLP:journals/corr/LongSD14}}&\textbf{84.7\%}  &37.2\%  \\
			\hline
			\textbf{Ours methods:}\\
			Baseline   &80.9\%   &32.0\%  \\
			Data Flipping &80.8\%  &36.3\%  \\
			Hierarchical Augmentation  &80.7\%  &40.1\%  \\
			Full Model  &82.0\%  &\textbf{41.1\%}  \\
			\hline
			
		\end{tabular}\label{sift_table}
	\end{table}

	In addition to the above comparison with other approaches, we discuss the effects of different steps for the proposed method. Obviously, the mean-class accuracy of the original baseline is not very high. But after the traditional data flipping which augment the training set, the performance increases from 32\% to 36.3\%. The incensement becomes larger with the hierarchical data augmentation for a more balanced data augmentation (40.1\%) and with a soft restricted MRF optimization for a smoother labeling (41.1\%). These statistics give a hint that the proposed method is more effective than the competitors.

\begin{figure*}
		\centering
		\includegraphics[width=0.95\textwidth]{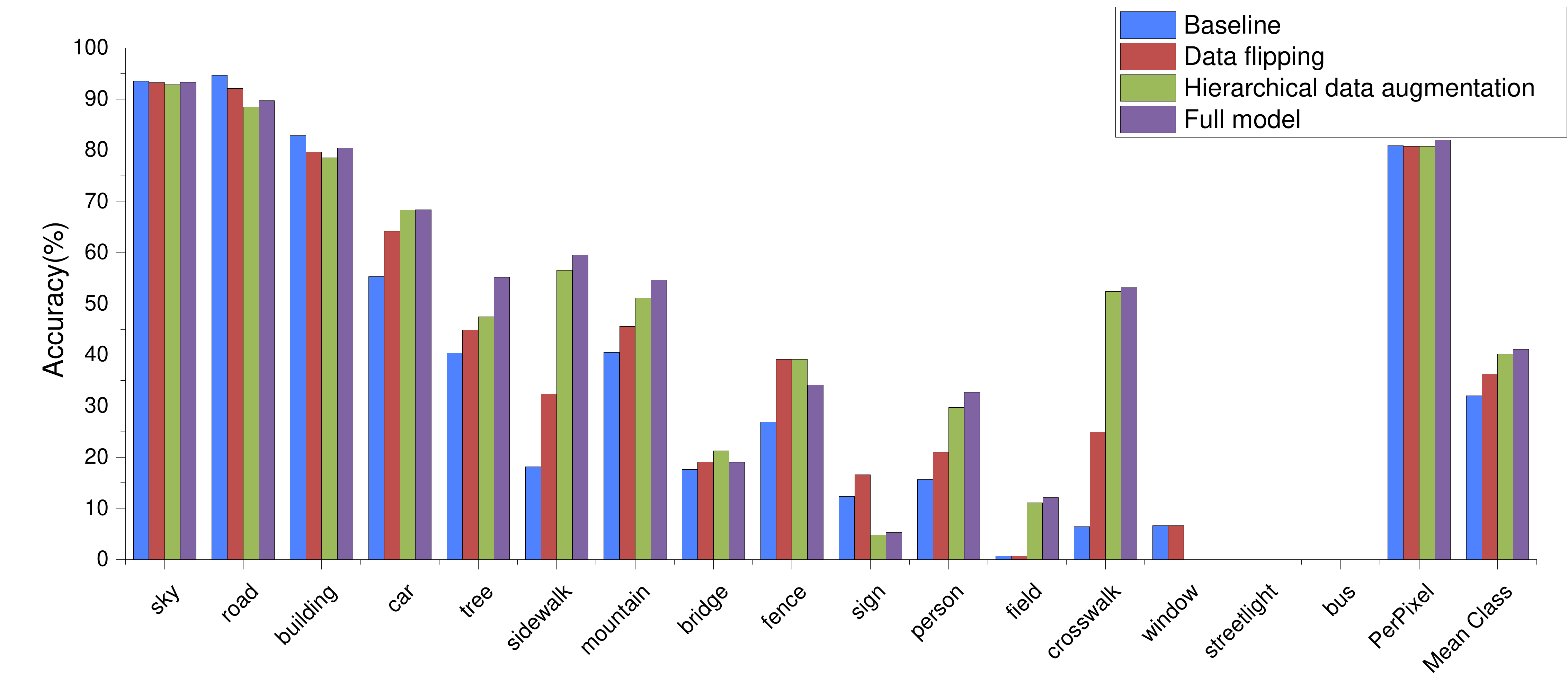}
		\caption{The performance of each class and two metrics in the four stages. }\label{Fig-result}
	\end{figure*}

	Fig. \ref{Fig-result} illustrates the per-pixel accuracy, overall per-pixel accuracy and mean-class accuracy. The data statistics are similar to that of  CamVid dataset: the performance of background objects decrease a little, and many foreground objects are promoted dramatically. The significant improvement of foreground objects labeling benefits from our priori s-CNNs trained on more foreground data after the hierarchical data augmentation. However, some foreground objects are not labeled correctly, such as ``streetlight'', ``bus'' and ``window''. The ``streetlight'' is similar to the ``pole'' in CamVid dataset, so it can not be segmented effectively and augmented. And the ``bus'' can not be trained enough because the training data are so rare that the effort of data augmentation is limited. The ``window'' is misclassified as ``building'' in the labeling stage.

\begin{figure*}
		\centering
		\includegraphics[width=.95\textwidth]{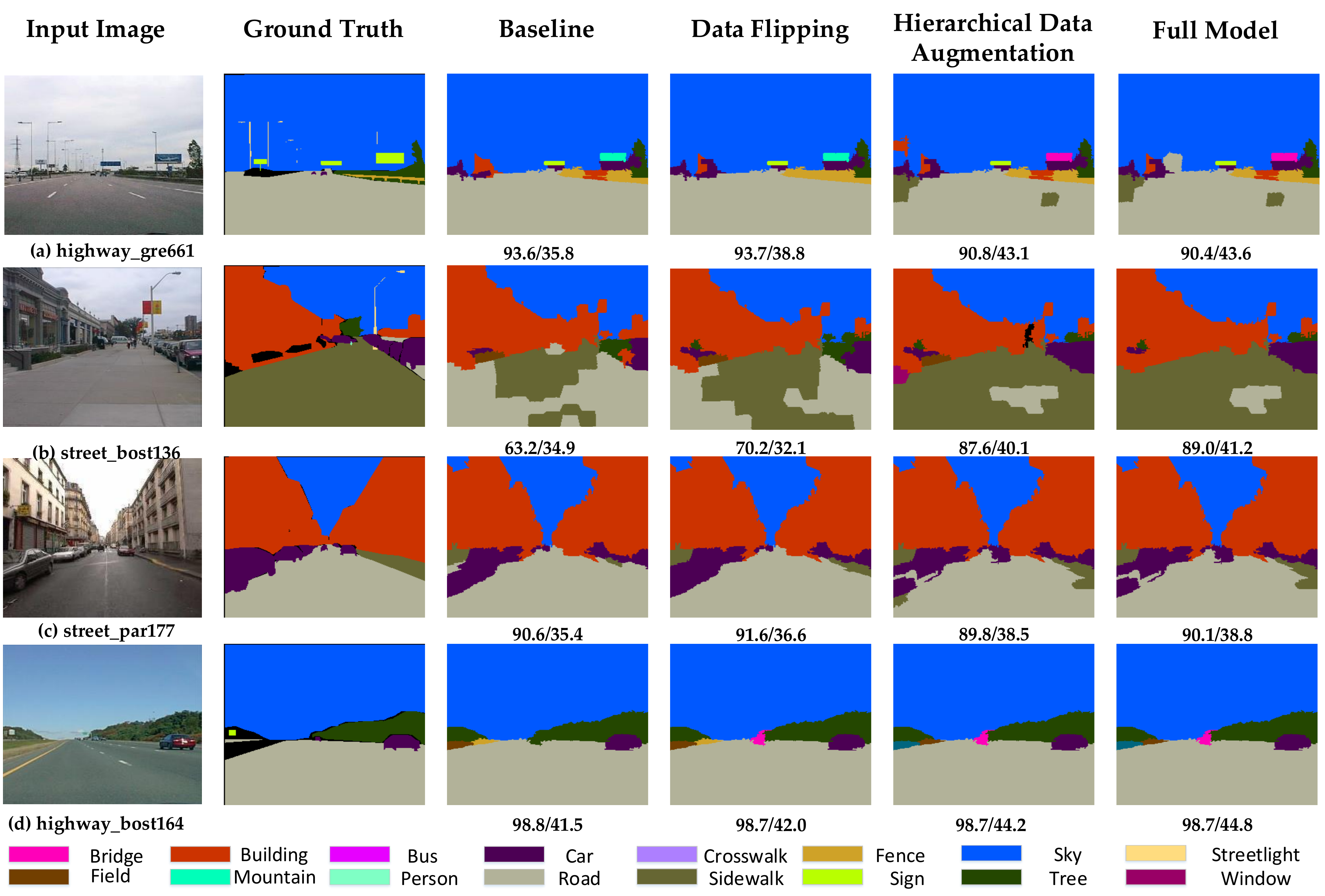}
		\caption{Exemplar results from SIFT Flow Street dataset. The value under the each image labeled is the percentage of per-pixel and mean-class accuracy respectively. }\label{resultshow}
	\end{figure*}

	Four typical results are shown in Fig. \ref{resultshow} to explain the effects of the hierarchical data augmentation and the soft restricted MRF inference. From (b) and (c), the ``sidewalk'' can be labeled more accurately with the hierarchical data augmentation than the baseline and traditional data flipping. In (c), the car in the right road is mislabeled as ``road'' by the first two methods, but our proposed augmentation method can label it correctly. In addition to the above intuitive displays, the statistics under the labeling images also illustrate the advantages of our augmentation strategy: the mean-class accuracy of each of the above images is promoted significantly by our data augmentation. As for the soft restricted MRF inference, in (b), the region mislabeled as ``road'' sidewalk shrink significantly because of its adjacent superpixels' smoothness. Similarly, some noises in the left (a) and (c) are reduced by contextual smoothing.
	
	\subsection{Effect of Prior s-CNNs}
	\label{Prior_s-CNNs}

\begin{figure}
		\centering  
		\subfigure[CamVid dataset.] { \label{fig_loss_cam} 
			\includegraphics[width=0.46 \columnwidth]{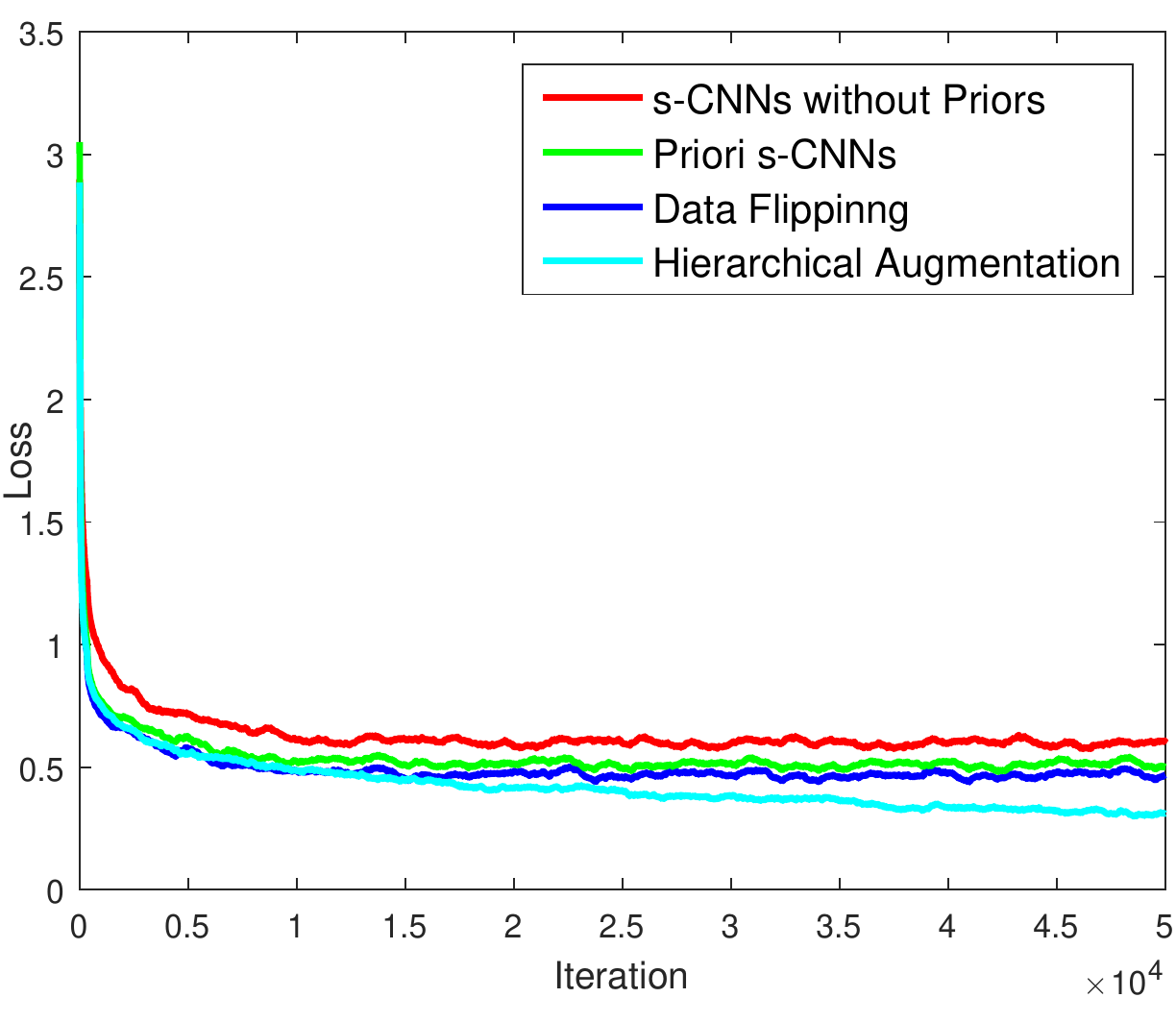} }    
		\subfigure[SIFT Flow Street dataset.] { \label{fig_loss_sift}    
			\includegraphics[width=0.46 \columnwidth]{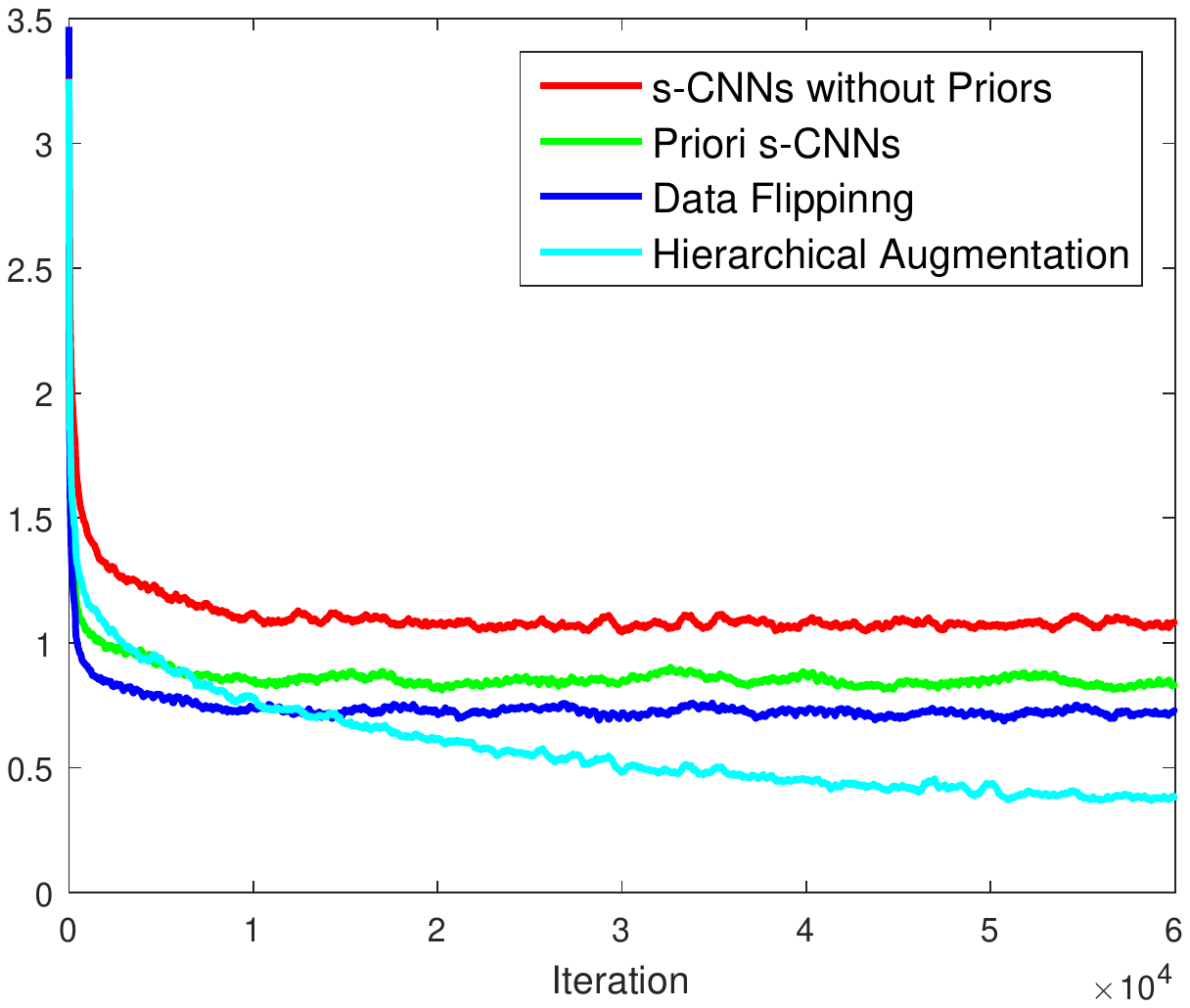} }    
		\caption{Convergence curves of the stepwise models: s-CNNs without priors, priori s-CNNs, data flipping CNNs and hierarchical augmentation CNNs. The left and right present the results on CamVid and SIFT Flow Street dataset, respectively.}    \label{fig_loss}  
	\end{figure}

	In the Section \mbox{\ref{psCNNs}}, the learned location priors are explained theoretically. In order to show the effect of priori s-CNNs intuitively, the verified experiments are added. For comparison, the s-CNNs without location priors are trained on the two datasets. To be specific, during the training and testing stages, each superpixel is shifted by a random value at the $x$ and $y$ axes respectively in the original image, which removes the location priors from the superpixels input. In practice, the input superpixel image, the shift values $\Delta x,\Delta y \in [ - 255,255]$ (image size is $256 \times 256$) are generated randomly on the X and Y-axis, respectively. If the superpixel is moved to the outside of the image, the $\Delta x$ and $\Delta y$ are regenerated until the new location of the superpixel is still in the image. That way, in an input image, all superpixels are move to a different random location, which eliminates the location priors in the original image. And the s-CNNs focuses on learning the features from the appearance information. The quantitative results are shown as in Table \mbox{\ref{priors_table}}. Specifically, the performance of the proposed priori s-CNNs is superior to that of s-CNNs without priors on the two datasets, which verifies the effectiveness of the former. In addition, the convergence curves of both models during training stage are shown in Fig. \mbox{\ref{fig_loss}}. Obviously, the priori s-CNNs converges a lower loss value than s-CNNs without priors.
	
	\begin{table}[htbp]
		
		\centering
		\caption{Comparison of s-CNNs without priors and priori s-CNNs on the two datasets.}
		
		\begin{tabular}{lcccccc}
			\hline
			Methods &Per-pixel &Mean-class \\
			\hline
			\textbf{CamVid dataset}\\
			s-CNNs without priors   & 70.6\% & 35.2\% \\
			priori s-CNNs (Baseline) & \textbf{77.1\%} & \textbf{45.6\%} \\
			\hline
			\textbf{SIFT Flow Street dataset} \\
			s-CNNs without priors   & 79.8\% & 25.8\%\\
			priori s-CNNs (Baseline)   &\textbf{80.9\%}   &\textbf{32.0\%}  \\
			\hline
			
		\end{tabular}\label{priors_table}
	\end{table}

	\subsection{Effect of Soft Restricted MRF}

	\label{soft_MRF}
\begin{figure}
		\centering  
		\subfigure[CamVid dataset.] { \label{fig_soft_cam} 
			\includegraphics[width=0.46 \columnwidth]{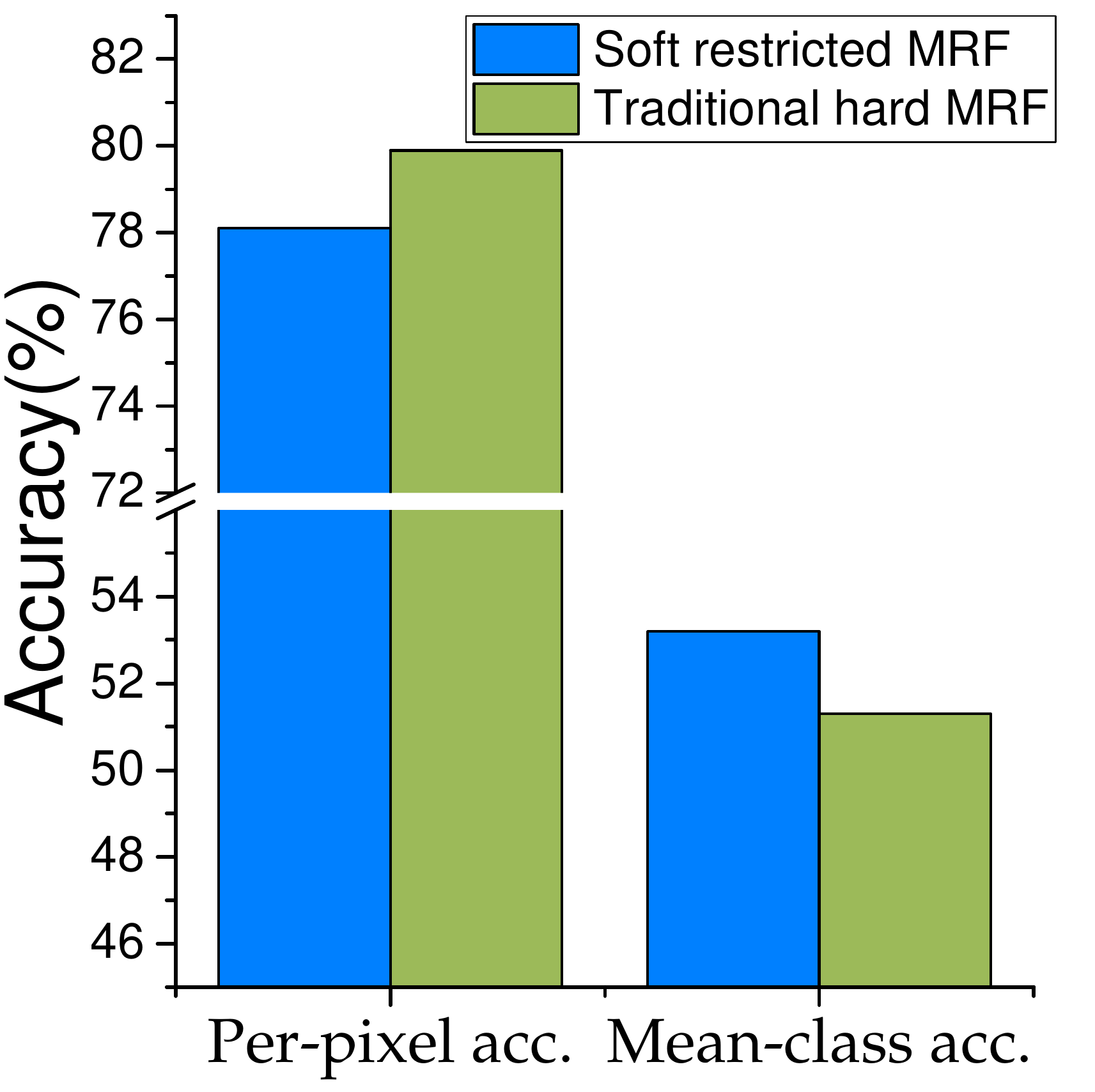} }    
		\subfigure[SIFT Flow Street dataset.] { \label{fig_soft_sift}    
			\includegraphics[width=0.46 \columnwidth]{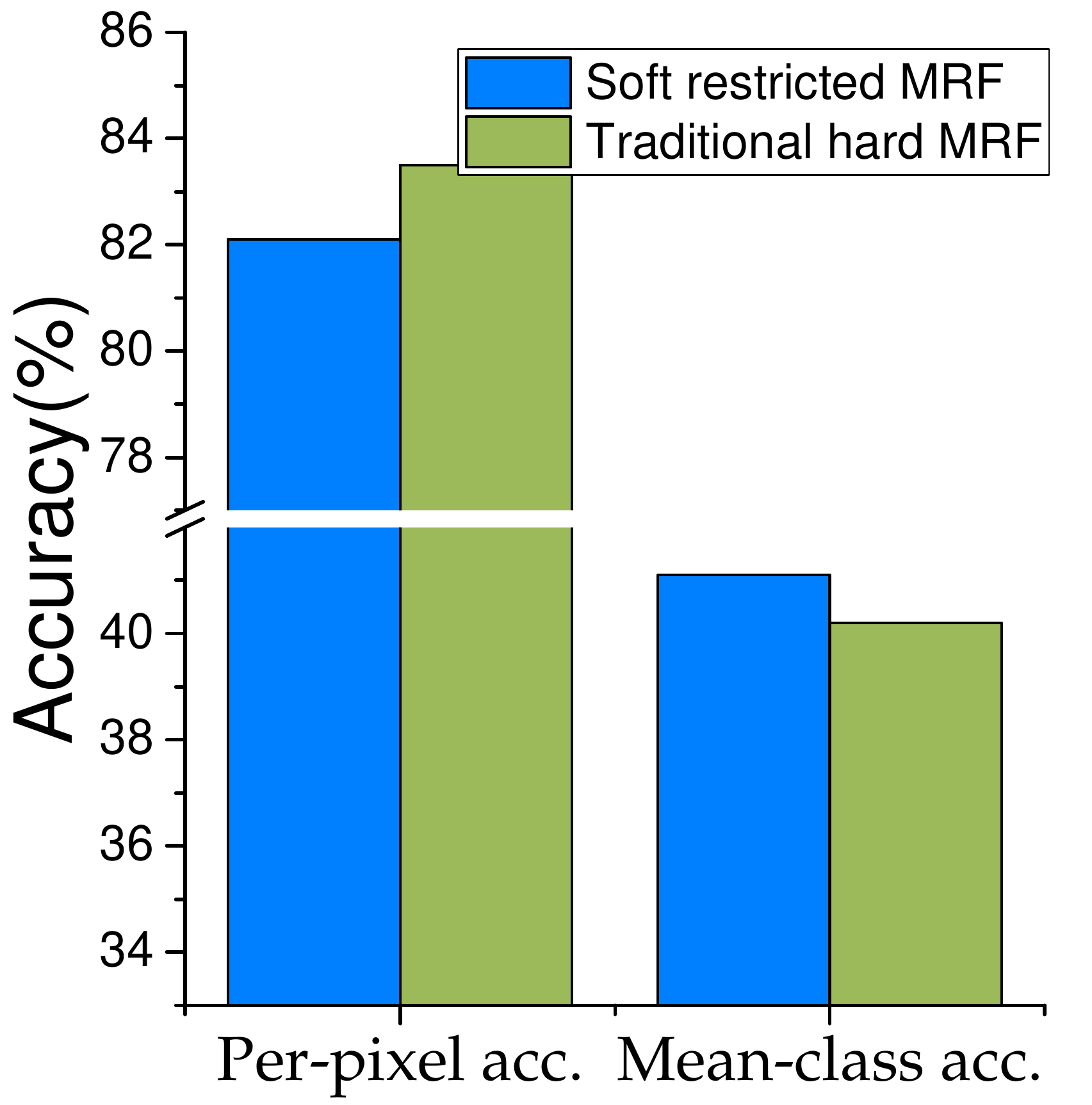} }    
		\caption{Comparison of our proposed soft restricted MRF and traditional hard MRF at the optimal $lamda$ $0.5$. }    \label{soft_all}  
	\end{figure}

	For explaining the advantage of the proposed soft restricted MRF intuitively, the comparisons with traditional hard MRF  \cite{DBLP:journals/ijcv/TigheL13} on the two datasets are illustrated in Fig. \mbox{\ref{soft_all}}. As can be seen from the bar chart, our method obtains higher mean-class accuracy than the traditional hard MRF, while the traditional method achieves a better per-pixel accuracy. But for the intelligent driving application, traditional hard MRF is not a good strategy, which sacrifices the performance of foreground labeling to get more overall per-pixel accuracy, because the foreground objects are more essential to safe driving than backgrounds. Therefore, the mean-class accuracy is more important than per-pixel accuracy. From this perspective, our model is much superior to the traditional model.
	
	\subsection{Comparison of CNNs v.s. Hand-crafted Features for Image Retrieval}
	\label{compare_features}
	In the $k$ nearest image retrieval, the deep features are exploited instead of the hand-crafted global features, such as spatial pyramids, GIST and RGB-color histograms in SuperParsing \mbox{\cite{DBLP:journals/ijcv/TigheL13}}. Because of classification capability of AlexNet, the 4,096-D feature in fc7 layer can represent the appearance and semantic information better than traditional features. Thus, the more accurate contextual information will be transferred to test images. 
	
	\begin{figure*}
		\centering
		\includegraphics[width=0.98\textwidth]{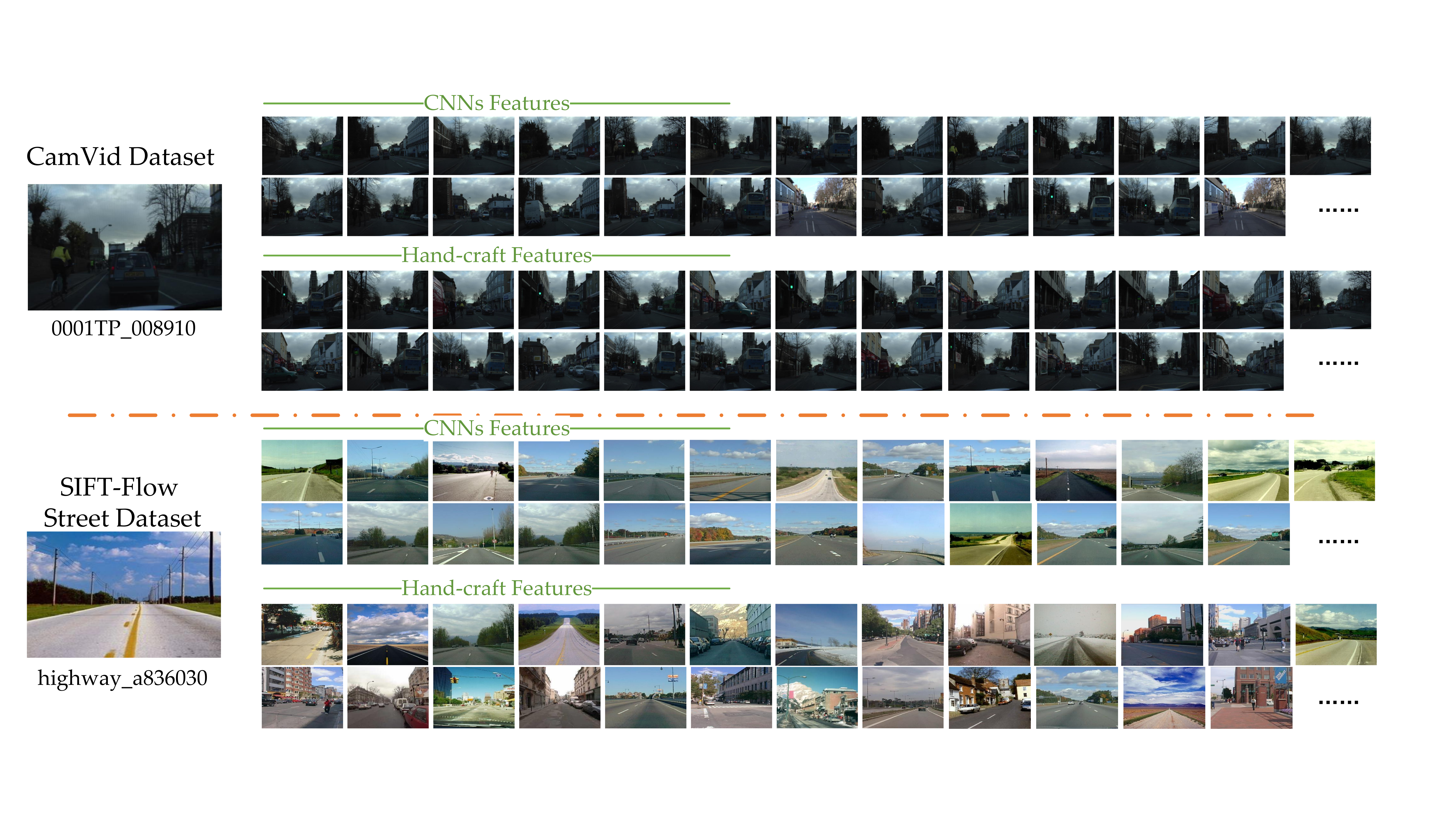}
		\caption{The exemplar display of the retrieval sets generated by different global image features (CNNs versus hand-crafted features). }\label{Fig-knn}
	\end{figure*}

	In order to show advantages of deep features, we exploit two types of features to find similar images and transfer contextual information. Table \mbox{\ref{CNN_feature_table}} shows the results of the two types of features in the full model. From the results, the improvement is not significant on CamVid dataset. To be specific, the retrieval results by two types of features are not so different. The main reason is that CamVid dataset includes continuous and similar image sequence and the retrieval set can be easily and accurately searched by hand-crafted features. On the SIFT Flow Street dataset, the improvement is obvious because of various scenes in the dataset so that the CNNs features demonstrate significant superiority.

	\begin{table}[htbp]
		
		\centering
		\caption{Comparison of CNNs v.s. hand-crafted features for image retrieval.}
		
		\begin{tabular}{lcccccc}
			\hline
			Methods &Per-pixel &Mean-class \\
			\hline
			\textbf{CamVid dataset}\\
			Hand-crafted features   &77.7\%  &53.0\%  \\
			CNNs features &\textbf{78.1\%}  &\textbf{53.2\%}  \\
			\hline
			\textbf{SIFT Flow Street dataset} \\
			Hand-crafted features   &81.4\%  &40.2\%  \\
			CNNs features &\textbf{82.0\%}  &\textbf{41.1\%}  \\
			\hline
			
		\end{tabular}\label{CNN_feature_table}
	\end{table}
	
	For showing the difference of the two types of features intuitively, we select two typical test images from the two datasets and display respectively the retrieval results in Fig. \mbox{\ref{Fig-knn}}. The left larger images are the query samples, and the right small images are the retrieval sets. Because of the limited space, the top-25 retrieved images are only displayed. About the ``0001TP\_008910'' image in CamVid dataset, although the results of the two methods are similar as a whole, there are some subtle difference. The hand-crafted features are so sensitive to the color information that they ignore the images from other scenes. However, the CNNs features' results include some images that have the same content with the test image from other scenes. As for the test image ``highway\_1836030'' in SIFT Flow Street dataset, the KNN that exploits CNNs features finds more similar images than the KNN that adopts three hand-crafted features. This is because the CNNs features describe more higher-level image representation including appearance, contextual and structural information than traditional hand-crafted features.

	\subsection{Convergence Analysis of the Stepwise Models}
	\label{convergence}
	In this Section, we show the convergence curves of each stepwise models during the training stage, namely s-CNNs without priors, priori s-CNNs, data flipping CNNs and hierarchical augmentation CNNs in Fig. \mbox{\ref{fig_loss}}. As can be seen from the loss curves, the first three models converge after around $20,000$ iterations. The last model on the two datasets, however, converge at about $50,000$ and $60,000$ iterations, respectively. The main reasons are aggravating imprecise segmentation noises and increasing training samples caused by the hierarchical data augmentation. But eventually, the hierarchical data augmentation CNNs converges a lower training loss value and achieves a higher classification performance on the test set than the former three.

	\section{Conclusion and Future Work}
	\label{conclusion}
	This paper proposes a joint framework of priori s-CNNs and soft restricted context transfer for street scenes labeling. The priori s-CNNs can fully exploit priori information through preserving superpixels' location in the image. Besides, it learns rich and discriminative features by the proposed hierarchical data augmentation. Compared with the traditional equal and random data augmentation, the proposed strategy can not only improve foreground objects labeling and mean-class accuracy significantly but also maintain the background objects labeling performance at the high level. In the context transfer, our proposed soft restriction on the smooth term of the MRF energy function can effectively reduce over smoothness, which makes the foreground objects not be improperly smoothed by the adjacent background objects. Extensive experiments have verified the effectiveness of the proposed method on the street scene datasets. Not limited to these street scenes, the proposed method also applies to other scenes (such as indoor and clothing parsing scenes) because no specific scene constraints are supposed in our approach.

	With the proposed framework, the labeling accuracy of the foreground objects increases significantly. Nevertheless, the missing and false labeling phenomena are common in our results. Thus, we will focus on integrating objects detector into our model to enhance the labeling accuracy in the future.

	\bibliographystyle{IEEEtran}
	\bibliography{IEEEabrv,reference}

	\begin{IEEEbiography}[{\includegraphics[width=1in,height=1.25in,clip,keepaspectratio]{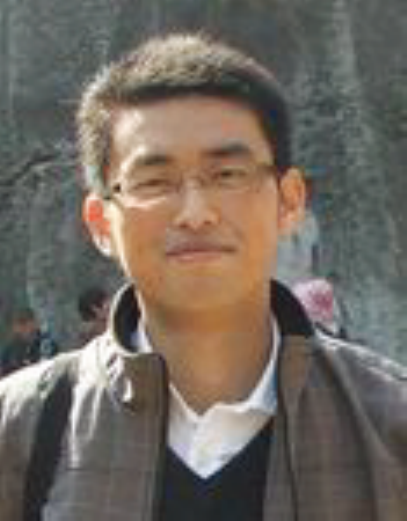}}]{Qi Wang} (M'15-SM'15) received the B.E. degree in automation and Ph.D. degree in pattern recognition and intelligent system from the University of Science and Technology of China, Hefei, China, in 2005 and 2010 respectively. He is currently a Professor with the School of Computer Science, with the Unmanned System Research Institute, and with the Center for OPTical IMagery Analysis and Learning, Northwestern Polytechnical University, Xi'an, China. His research interests include computer vision and pattern recognition.
	\end{IEEEbiography}
	
	\begin{IEEEbiography}[{\includegraphics[width=1in,height=1.25in,clip,keepaspectratio]{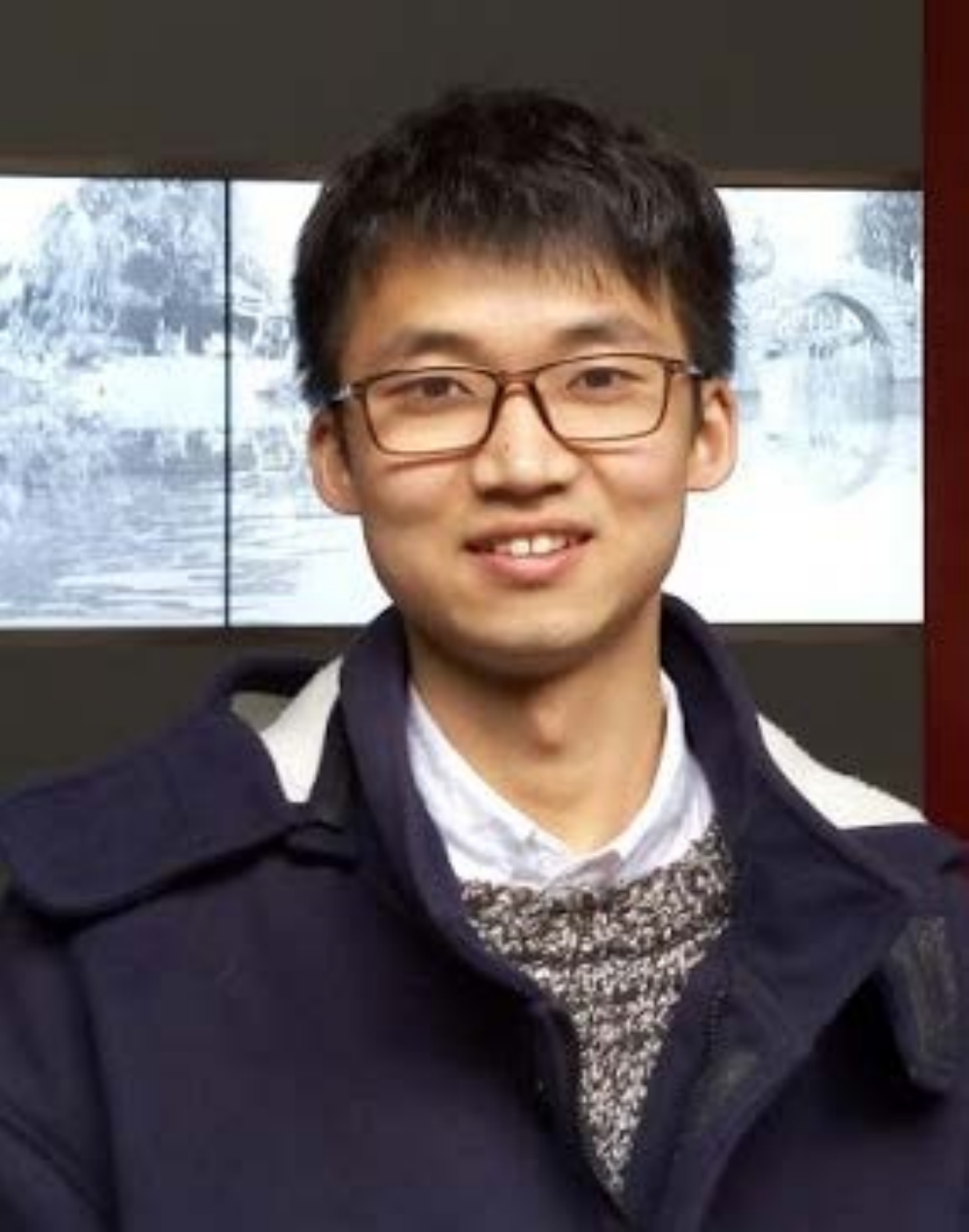}}]{Junyu Gao} received the B.E. degree in computer science and technology from the Northwestern Polytechnical University, Xi'an 710072, Shaanxi, P. R. China, in 2015. He is currently pursuing the Master degree from Center for Optical Imagery Analysis and Learning, Northwestern Polytechnical University, Xi’an, China. His research interests include computer vision and pattern recognition.
	\end{IEEEbiography}
	
	\begin{IEEEbiography}[{\includegraphics[width=1in,height=1.25in,clip,keepaspectratio]{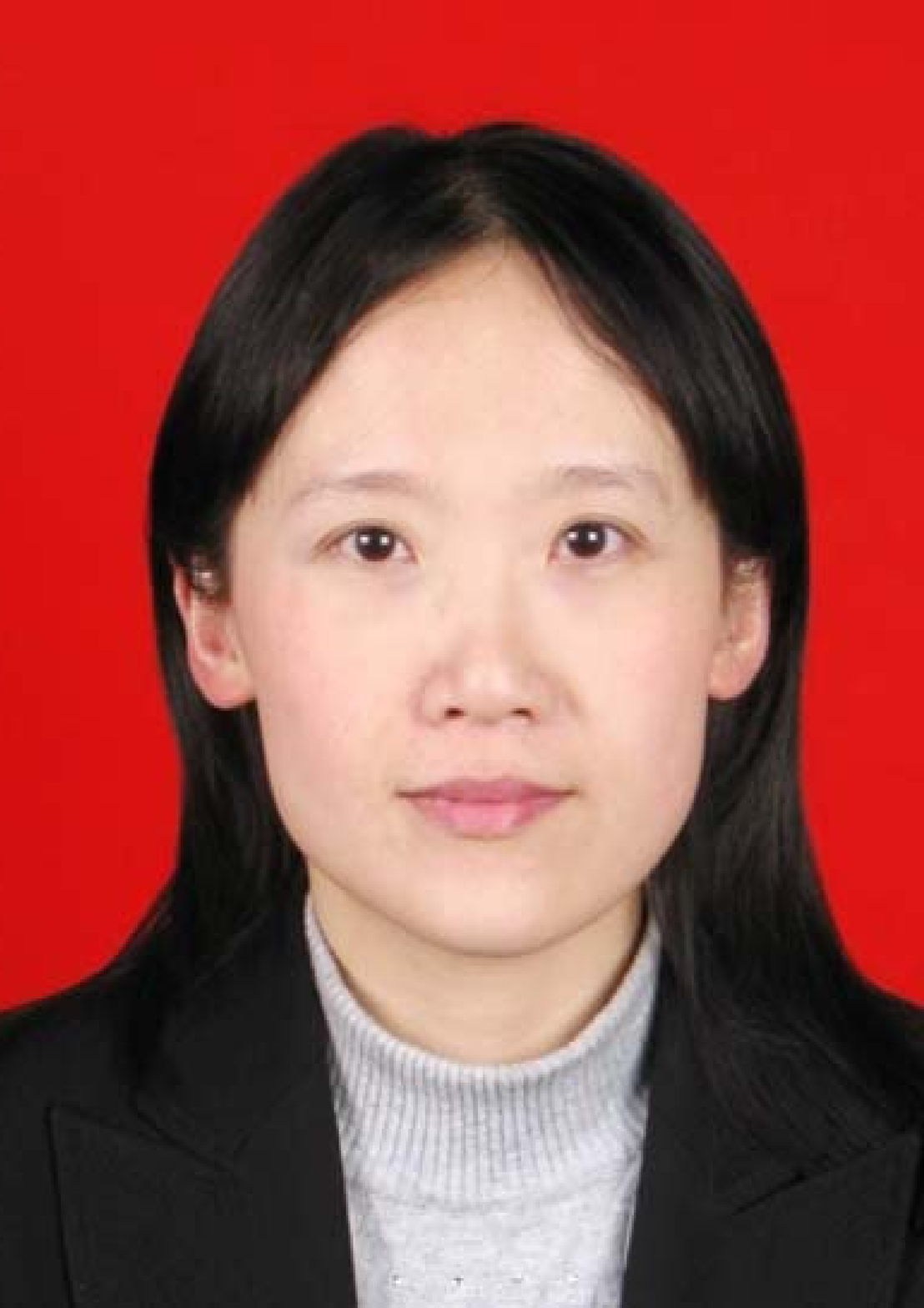}}]{Yuan Yuan} (M'05-SM'09) is currently a full professor with the School of Computer Science and Center for OPTical IMagery Analysis and Learning (OPTIMAL), Northwestern Polytechnical University, Xi'an 710072, Shaanxi, P. R. China. She has authored or coauthored over 150 papers, including about 100 in reputable journals such as IEEE Transactions and Pattern Recognition, as well as conference papers in CVPR, BMVC, ICIP, and ICASSP. Her current research interests include visual information processing and image/video content analysis.
	\end{IEEEbiography}

\end{document}